\PassOptionsToPackage{table}{xcolor}

\documentclass[manuscript]{acmart}

\usepackage{amsmath,amsfonts}
\usepackage{array}
\usepackage{subcaption}
\usepackage{multirow}
\usepackage{tabularx}
\usepackage{colortbl}
\usepackage{threeparttable}
\usepackage{tikz}
\usetikzlibrary{positioning,arrows.meta}

\newcolumntype{Y}{>{\centering\arraybackslash}p{0.3\columnwidth}} 

\providecommand{\stat}[1]{#1}

\usepackage{soul}
\definecolor{revisionhl}{HTML}{FFF200}
\sethlcolor{revisionhl}
\newcommand{\revised}[1]{\hl{#1}}
\renewcommand{\revised}[1]{#1}

\setcopyright{acmlicensed}
\copyrightyear{2026}
\acmYear{2026}
\acmDOI{XXXXXXX.XXXXXXX}
\acmJournal{TIST}
\acmVolume{0}
\acmNumber{0}
\acmArticle{0}
\acmMonth{0}

\begin{document}

\title{Modeling Event-Conditioned Emotional Expression Response with LLM
  Agents in a Synthetic Social Simulation}

\author{Rebecca Marigliano}
\correspondingauthor
\orcid{0009-0003-8271-9990}
\email{rmarigli@andrew.cmu.edu}
\affiliation{%
  \institution{Carnegie Mellon University}
  \city{Pittsburgh}
  \state{Pennsylvania}
  \country{USA}
}

\author{Peter Carragher}
\orcid{0000-0003-4053-1001}
\email{petercarragher@cmu.edu}
\affiliation{%
  \institution{Carnegie Mellon University}
  \city{Pittsburgh}
  \state{Pennsylvania}
  \country{USA}
}

\author{Kathleen M. Carley}
\orcid{0000-0002-6356-0238}
\email{kathleen.carley@cs.cmu.edu}
\affiliation{%
  \institution{Carnegie Mellon University}
  \city{Pittsburgh}
  \state{Pennsylvania}
  \country{USA}
}

\renewcommand{\shortauthors}{Marigliano et al.}

\begin{abstract}
Emotions are central drivers of human behavior, yet most behavioral models fail to capture their role in shaping perception, decision-making, and collective action. Application areas such as emergency response and public health are especially constrained by the lack of emotion-aware modeling. While large language models (LLMs) are increasingly integrated into agent-based models (ABMs), little work has examined how to endow LLM agents with emotional reasoning or validated their ability to reproduce empirically observed emotional patterns. This study addresses this gap by introducing a hierarchical prompting framework that conditions LLM agents with demographic, psychological, and contextual cues to generate event-conditioned emotional responses to real-world events. We validate the resulting event-conditioned emotional responses against large-scale social media reaction data derived from the Social Science One Facebook URL Shares Dataset \cite{socialscienceone2023urls}, complemented by demographic and personality benchmarks. Our simulation results reproduce both polarized and consensus-driven emotional patterns with significant fidelity,  supporting the application of LLM agents for emotion-aware modeling in behavioral simulations. We also find, however, that systematic gaps in emotional expression of LLM agents and human populations exist, such as the underexpression of anger relative to human baselines. As such, these findings highlight the need for continued calibration if LLM agents are to serve as proxies for human populations in behavioral modeling. By quantifying their divergence from human populations in terms of emotional expression, we hope that our methods can aid in this effort.
\end{abstract}



\maketitle

\section{Introduction}
Emotional reasoning is a central driver of human behavior. In March 2020, COVID-19 caused an 891\% surge in calls to the federal Disaster Distress Helpline compared to the previous year \cite{abcnews2020distress}. The helpline was criticized for failing to anticipate both the scale of public distress and how emotions such as fear, isolation, grief, and anger spread through social systems. During the same period, more than 40\% of Americans reported adverse mental health conditions \cite{czeisler2020mental}, overwhelming existing support systems. Accounting for emotional amplification and diffusion could have enabled more effective planning and targeted interventions.

Similar dynamics occur in online misinformation and persuasion. Dampening affective responses can reduce engagement with misinformation that exploits fast, emotion-driven reasoning rather than deliberative thought \cite{kahneman2011thinking,evans2008dual,stanovich2000individual}. Conversely, emotional appeals can be used to promote credible information, as demonstrated in public health messaging during the COVID-19 pandemic \cite{van2020using}.

Modeling emotional response is therefore important for policy design and behavioral simulation. Emotional reasoning influences risk perception, decision making, and collective behavior \cite{kahneman2011thinking}. Emotional contagion contributes to panic and polarization, while trust, empathy, and cooperation depend on emotional reciprocity \cite{sapolsky2017behave}. Models that ignore emotional reasoning omit an important component of social behavior and opinion dynamics \cite{loewenstein2000emotions}, with evidence showing it is as important as deliberative reasoning in these settings \cite{slovic2007affect}.



Despite this, traditional models of behavior and opinion dynamics emphasize belief updating over emotional response. To address this limitation, we make three contributions. First, we introduce a hierarchical framework for generating LLM personas capable of producing event-conditioned emotional responses across country, community, and individual levels. Second, we formalize an event-conditioned emotional response model linking trait-level affect, event salience, and stance to expressed emotion. Third, we evaluate the framework using Facebook reaction data from the overturning of Roe v.~Wade and the expansion of the 988 Suicide Hotline during the Covid pandemic, assessing whether synthetic personas reproduce observed demographic differences in emotional expression. 

Our framework achieves this by conditioning personas on demographic information, moral foundations, OCEAN personality traits, ideological context, and event-specific information through structured multi-stage prompting. The resulting personas engage in repeated network-mediated interactions through a shared communication environment, allowing emotional responses to evolve across successive interaction timesteps.

We evaluate the framework through both internal and external validation. Internally, ablation experiments assess response quality using measures of emotional intensity, topical relevance, and persona consistency, together with implementation-level validation of demographic fidelity, hierarchical consistency, opinion generation, and schema integrity (Appendix~\ref{appendix:internal_consistency}). Externally, generated emotional response patterns are compared with Facebook URL Shares engagement data, demonstrating that the personas reproduce several demographic and event-conditioned trends observed in real-world discourse \cite{socialscienceone2023urls}. Supplementary COVID-19 experiments further extend this evaluation to prolonged and politically polarized public-health discourse. The resulting demographic and emotional trends are consistent with prior findings on age-related emotional regulation \cite{wrosch2017anger_sadness_control}.

While our results demonstrate the potential of LLM-based synthetic social simulation for emotional modeling at the population level, they do not establish a psychological equivalence between LLM personas and human subjects, nor do they seek to diminish the importance of human subjects in behavioral research. More broadly, the findings contribute to ongoing discussions of whether LLM-generated personas can support meaningful social simulation in resource constrained conditions \cite{davidson_integrating_2025}. Agreement with observed age-, anger-, and sadness-related trends supports applications such as ``Silicon Sampling'' \cite{liu_improving_2023,park_generative_nodate,binz_centaur_2024,chopra2023agenttorch}, while remaining gaps, including the underrepresentation of anger, highlight the need for continued empirical calibration \cite{van_rooij_reclaiming_2024,schroder_large_2025}. As such, we hope that this work clarifies the positive use cases of LLM personas for emotional and behavioral modeling, while also serving as a foundation for their continued calibration in synthetic social simulations.

\section{Related Works}
\label{sec:related_works}

Emotional response in political contexts is motivated by persistent findings that emotional content spreads rapidly on social media \cite{bail2018exposure}, with anger showing particularly strong viral characteristics \cite{brady2017mad}, and emotional contagion being especially pronounced in digital settings \cite{kramer2014experimental}. Computational approaches have evolved from early emotion theories and sentiment polarity detection to richer emotional profiling systems grounded in psychological theory and supported by lexicons and modern language models.

\paragraph{Cross-Cultural Modeling}
Plutchik’s theory of primary emotions \cite{plutchik1980general} underpins widely used computational resources such as the NRC Emotion Lexicon \cite{mohammad2013crowdsourcing}. While work has shown universality in some emotion signals (e.g., facial expressions) \cite{ekman1992argument}, cultural variation in recognition and expression norms remains well documented \cite{jack2012facial}, motivating calibration of emotion resources trained primarily on Western data \cite{mohammad2016sentiment}. To this end, cross-cultural research efforts have extended traditional western-oriented emotion lexicons by calibrating them for broader applicability \cite{li2023emotionalstimuli, tsai2000emotionaging}.


\paragraph{Moral Foundation Theory and Political Expression}
Moral Foundations Theory provides a structured account of cross-cultural and ideological variation in moral reasoning \cite{graham2009liberals}. Cultural dimensions and expression norms have been linked to systematic differences in affective expression across societies \cite{hofstede2001culture,matsumoto2008mapping}, including distinctions between American and Canadian discourse patterns \cite{lipset1990continental,adams2003fire}. Within the United States, ideological groups differ in moral-emphasis profiles \cite{graham2009liberals}, and perceived moral violations are associated with distinct emotional responses (e.g., sadness, anger, contempt) \cite{clifford2015moral,van2014moral}.

\paragraph{Personality Modeling}
Cross-cultural differences also appear in personality distributions \cite{schmitt2007geographic}. Big Five traits are associated with systematic variation in affective expression: extraversion correlates with positive affective expression and neuroticism predicts negative emotion intensity \cite{mccrae1992introduction,watson2000mood}. Individual differences in regulation further shape expression patterns, with neuroticism linked to maladaptive regulation and conscientiousness to more controlled expression \cite{gross2001emotion}.

\paragraph{LLM Personas}
Recent work suggests LLM-based agents can sustain coherent personas over extended interactions \cite{park2023generative} and can produce behaviorally plausible responses distributions when conditioned on persona specifications \cite{argyle2023out}. 
Prompting methods also provide finer control over model behavior by explicitly eliciting intermediate reasoning and structured outputs through staged, multi-step prompts \cite{wei2022chain,zhou2022least}. These advances motivate structured persona conditioning as a practical pathway for integrating richer psychological and contextual constraints into synthetic agent behavior.

\paragraph{Limitations}
Emotion expression varies by communication modality, with differences observed between text-based and vocal contexts \cite{bazarova2012self}. More broadly, LLM-generated personas may reflect biases in training data \cite{weidinger2021ethical,bender2021dangers}. In addition, some emotions may be systematically underrepresented in LLM outputs (e.g., attenuated anger), raising concerns about ecological validity for polarized scenarios. Finally, validation is often limited by topic coverage, constraining generalizability. These limitations reinforce the need for continued calibration and cross-platform validation before deploying LLM personas in high-stakes behavioral modeling.

\section{Method: Hierarchical Persona Synthesis}
\label{sec:methods}

Persona generation proceeds hierarchically: countries $\rightarrow$ communities $\rightarrow$ individuals. At each stage, structured prompts are passed to an LLM to produce records in a predefined schema. A validation layer checks completeness and demographic consistency; invalid generations are retried until valid outputs are obtained. This design helps maintain consistency between country-, community-, and individual-level attributes.
\begin{figure}[htbp]
    \centering
    \includegraphics[width=0.62\linewidth]{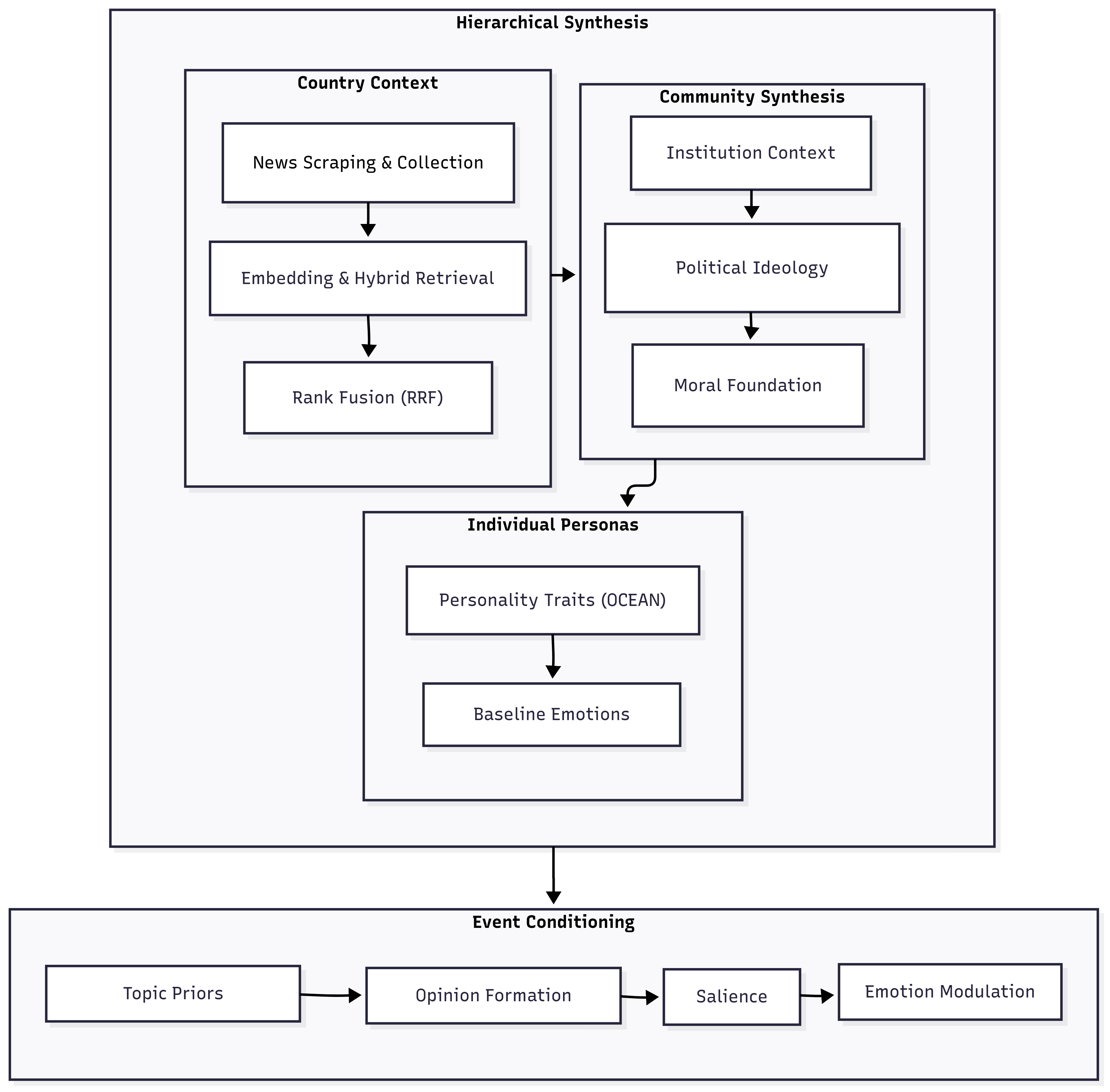}
    \caption{Persona construction and event-conditioned response pipeline. News grounding supports hierarchical synthesis, while traits, stance, and salience shape emotional responses.}
    \label{fig:pipeline}
    \Description{Flowchart of the persona construction pipeline. A
      Hierarchical Synthesis block contains a Country Context column (news
      scraping and collection, then embedding and hybrid retrieval, then rank
      fusion) feeding a Community Synthesis column (institution context, then
      political ideology, then moral foundation), which in turn feeds an
      Individual Personas block (personality traits, then baseline emotions).
      A separate Event Conditioning block below chains topic priors, opinion
      formation, salience, and emotion modulation.}
\end{figure}

A persona is represented as
\begin{equation}
P = f_{\text{persona}}(C, D_{\text{comm}}, \Psi, \Omega),
\end{equation}
where $P$ denotes a synthetic persona, $C$ represents country-level descriptors, $D_{\text{comm}}$ denotes community-level features, $\Psi$ contains psychological attributes (OCEAN traits, moral foundations), and $\Omega$ encodes event-conditioned parameters (stance, salience, emotion cues) (Figure~\ref{fig:pipeline}).

Table~\ref{tab:notation} summarizes the main notation used throughout the manuscript.

\begin{table}[htbp]
\centering
\caption{Summary of notation}
\label{tab:notation}
\footnotesize
\begin{tabularx}{\linewidth}{p{0.23\linewidth}X}
\toprule
\textbf{Symbol}& \textbf{Description} \\
\midrule
$P$ & Hierarchically synthesized persona.\\
$p$ & Persona index, $p \in \mathcal{P}$. \\
$\mathcal{P}$ & Set of synthetic personas. \\
$C$ & Country-level descriptors. \\
$D_{\text{comm}}$ & Community-level descriptors. \\
$\Psi$ & Psychological attributes (OCEAN and Moral Foundations). \\
$\Omega$ & Event-conditioned stance, salience, and emotion. \\
$\mathcal{D}$ & News document collection. \\
$\mathcal{D}_k$ & Top-$k$ retrieved documents. \\
$k$ & Number of retrieved documents. \\
$s_{\text{vec}}(q,d)$ & Dense retrieval similarity score. \\
$s_{\text{bm25}}(q,d)$ & BM25 lexical similarity score. \\
$K_{\text{RRF}}$ & RRF smoothing constant. \\
$\mathbf{o}$ & OCEAN personality vector. \\
$\mathbf{m}$ & Moral Foundations vector. \\
$\mathbf{W}$ & OCEAN-to-emotion projection matrix. \\
$\mathbf{e}_{\text{base}}$ & Baseline emotion profile. \\
$\mathbf{e}(t)$ & Event-conditioned emotion profile at time $t$. \\
$s(t)$ & Event stance at time $t$. \\
$u(t)$ & Event salience at time $t$. \\
$\alpha$ & Emotion modulation factor. \\
$L$ & Political leaning score. \\
$I(L)$ & Ideology mapping function. \\
$f_{\text{MFT}}$ & Ideology-to-moral-values mapping. \\
$C_t$ & Conversational context at time $t$. \\
$\mathcal{S}$ & Event scenarios. \\
$\mathcal{L}$ & Prompting conditions. \\
$T$ & Interaction timesteps. \\
$\mathcal{K}_r$ & Response keywords. \\
$\mathcal{K}_s$ & Scenario keywords. \\
$I_{\text{emo}}$ & Emotional intensity metric. \\
$R_{\text{topic}}$ & Topical relevance metric. \\
$PC_{\text{cons}}$ & Persona consistency metric. \\
$Q$ & Overall quality score. \\
$N_{\text{exp}}$ & Number of experimental observations. \\
\bottomrule
\end{tabularx}
\end{table}

\subsection{Country Level Grounding}
Country grounding defines broad descriptors around demographics, political climate, and cultural context. National statistics provide baselines (age, gender, education, ethnicity), while contextual documents (news, reports) embed contemporary conditions. Scenario configuration files specify event-conditioned sampling rules. Demographic and personality parameters remain unchanged, serving as constraints that limit how far narrative variation can move the persona away from its underlying predispositions.

The personas show country-conditioned demographic calibration. As shown in Table~\ref{tab:persona_summary}, age, gender, and education closely match the selected national reference proportions after rounding to integer persona counts. Older adults form the largest age group in both country samples (Canada: 37.4\%; U.S.: 39.7\%), while gender ratios closely match national references (Canada: 50.4\% female; U.S.: 50.9\% female). Education levels also align with the selected reference distributions, with national differences preserved between Canada and the United States. Political orientation is not census-calibrated; instead, it is derived from scenario and engagement-weighted assumptions. Both country samples lean left-of-center, with similar distributions across ideological categories.

\begin{table}[htbp]
\centering
\caption{Summary of country-conditioned synthetic persona dataset used in experiments, with comparison to national demographic statistics.}
\label{tab:persona_summary}
\scriptsize
\setlength{\tabcolsep}{3.5pt}
\begin{tabular}{lccccc}
\toprule
\multirow{2}{*}{\textbf{Attribute}} &
\multicolumn{2}{c}{\textbf{Canada}} &
\multicolumn{2}{c}{\textbf{United States}} &
\textbf{Total} \\
\cmidrule(lr){2-3} \cmidrule(lr){4-5}
 & Sample & Ref.\% & Sample & Ref.\% & Sample \\
\midrule
\textit{Total Personas} & 270 & -- & 267 & -- & 537 \\
\midrule
\textit{Age Distribution}$^a$ & & & & & \\
\hspace{3mm} 18--24 & 17 (6.3\%) & 6.4 & 16 (6.0\%) & 6.0 & 33 (6.1\%) \\
\hspace{3mm} 25--34 & 36 (13.3\%) & 13.3 & 40 (15.0\%) & 14.9 & 76 (14.2\%) \\
\hspace{3mm} 35--44 & 35 (13.0\%) & 12.8 & 33 (12.4\%) & 12.5 & 68 (12.7\%) \\
\hspace{3mm} 45--54 & 41 (15.2\%) & 15.2 & 35 (13.1\%) & 13.2 & 76 (14.2\%) \\
\hspace{3mm} 55--64 & 40 (14.8\%) & 14.9 & 37 (13.9\%) & 13.9 & 77 (14.3\%) \\
\hspace{3mm} 65+    & 101 (37.4\%) & 37.4 & 106 (39.7\%) & 39.5 & 207 (38.5\%) \\
\midrule
\textit{Gender}$^b$ & & & & & \\
\hspace{3mm} Female & 136 (50.4\%) & 50.4 & 136 (50.9\%) & 50.8 & 272 (50.7\%) \\
\hspace{3mm} Male   & 134 (49.6\%) & 49.6 & 131 (49.1\%) & 49.2 & 265 (49.3\%) \\
\midrule
\textit{Education}$^c$ & & & & & \\
\hspace{3mm} Bachelor’s+ & 71 (26.3\%) & 26.4 & 112 (41.9\%) & 42.0 & 183 (34.1\%) \\
\hspace{3mm} Technical/College & 71 (26.3\%) & 26.4 & 80 (30.0\%) & 30.0 & 151 (28.1\%) \\
\hspace{3mm} High School & 128 (47.4\%) & 47.2 & 75 (28.1\%) & 28.0 & 203 (37.8\%) \\
\midrule
\textit{Political Orientation}$^d$ & & & & & \\
\hspace{3mm} Left & 70 (25.9\%) & -- & 69 (25.8\%) & -- & 139 (25.9\%) \\
\hspace{3mm} Center Left & 76 (28.1\%) & -- & 74 (27.7\%) & -- & 150 (27.9\%) \\
\hspace{3mm} Center & 39 (14.4\%) & -- & 39 (14.6\%) & -- & 78 (14.5\%) \\
\hspace{3mm} Center Right & 62 (23.0\%) & -- & 62 (23.2\%) & -- & 124 (23.1\%) \\
\hspace{3mm} Right & 23 (8.5\%) & -- & 23 (8.6\%) & -- & 46 (8.6\%) \\
\bottomrule
\end{tabular}
\begin{tablenotes}
\scriptsize
\item[$^a$] Age groups are rounded to preserve the country-level sample totals while closely matching national reference proportions.
\item[$^b$] Gender closely approximates national reference ratios.
\item[$^c$] Education categories are aggregated to match the available national reference categories.
\item[$^d$] Political orientation is derived from scenario and engagement-weighted digital participation assumptions, not census data.
\end{tablenotes}
\end{table}

The hierarchical pipeline generated $537$ personas with balanced cross-national representation ($267$ US, $270$ Canada). The sample size was determined by a power analysis which indicated that approximately 250 personas per country is sufficient to detect medium-to-large effects in emotional responses.

\subsection{Community Level Grounding}
Six institutional categories evenly contributed to the sample population (16.4--16.8\% each), with communities consistently sized at 8--9 personas. The LLM produced organizational variation within categories, including Chambers of Commerce, student unions, and professional societies. This category balance was imposed by the scenario configuration.

Communities inherit country descriptors and introduce meso-level variation (demographics, education, political leaning). Communities are created by combining national priors with issues of local importance retrieved through the news retrieval system.

\paragraph{News Retrieval for Event-Conditioned Prompting}
To ground persona generation in the event context, we retrieve contemporaneous domestic news as conditioning evidence for country-, community-, and individual-level synthesis. Given a news collection $\mathcal{D}$ and an event query $q$, dense semantic retrieval encodes the query and documents using the SBERT \texttt{all-MiniLM-L6-v2} sentence embedding model \cite{reimers2019sentence}.

\begin{align}
\mathcal{D}_k
&= \underset{d \in \mathcal{D}}{\operatorname{Top-k}}
   \bigl\{\operatorname{RRF}(d)\bigr\},\\
\operatorname{RRF}(d)
&= \sum_{r \in \{\text{vec},\,\text{BM25}\}}
   \frac{1}{K_{\text{RRF}}+\operatorname{rank}_r(d)},
\end{align}

where $K_{\text{RRF}}=60$ is the fusion constant. The retrieved documents provide factual context that conditions the generated personas' reasoning, moral framing, and emotional responses..

\paragraph{Ideological Anchoring via Moral Foundations}
Political leaning is represented by a continuous score $L\in[-1,1]$, which is mapped to an ideological category and subsequently to a Moral Foundations profile:

\[
I:[-1,1]\rightarrow
\{\text{Left},\text{Center Left},\text{Center},
\text{Center Right},\text{Right}\},
\]

\[
f_{\text{MFT}}(I(L))
\rightarrow
\mathbf{m}=
\begin{bmatrix}
m_{\text{care}},m_{\text{fairness}},m_{\text{loyalty}},\\
m_{\text{authority}},m_{\text{sanctity}},m_{\text{liberty}}
\end{bmatrix}^{\!\top}.
\]

Personas inherit ideological orientation from their institutional community. Political leaning $L$ is sampled from community-specific distributions ($\sigma=0.1$) and mapped to

\begin{equation}
\mathbf{m} = f_{\text{MFT}}(I(L)) + \boldsymbol{\varepsilon}, \qquad \boldsymbol{\varepsilon} \sim \mathcal{N}(\mathbf{0}, \sigma^2 \mathbf{I}),
\end{equation}

where $f_{\text{MFT}}$ assigns six Moral Foundations scores using five ordinal levels (Very Low--Very High; Appendix~\ref{app:moral}), and $\boldsymbol{\varepsilon}$ introduces bounded individual variation. Liberal personas receive higher Care and Fairness scores, whereas conservative personas receive higher Authority and Sanctity scores, with centrist ideologies occupying intermediate profiles.

\subsection{Individual Level Grounding}
Individuals are sampled within communities and enriched with psychological traits (personality, moral foundations, baseline emotions) to capture within-community variability.

\paragraph{Personality Trait Sampling}
Following the methodology introduced in the AURORA persona framework \cite{marigliano2025aurora}, each persona is assigned Big Five (OCEAN) personality traits sampled from country-specific Gaussian priors estimated from cross-cultural studies \cite{schmitt2007geographic}:
\begin{equation}
T_i \sim \mathcal{N}(\mu_{i,\text{ctry}}, \sigma_{i,\text{ctry}}^2), \qquad
T_i \in [0,1], \quad i \in {O,C,E,A,N}.
\end{equation}
In this work, we build on AURORA by using these personality traits as the basis for event-conditioned emotional modeling. Specifically, the sampled traits are combined with event salience and stance to generate emotional responses, which are then propagated through repeated network-mediated interactions and evaluated against real-world emotional engagement data.

\subsection{Quality Control}

Each LLM output is parsed and schema-validated; invalid records trigger retries. Batches proceed to analysis only if $\geq 70\%$ of outputs are valid, improving reproducibility and reducing spurious generations. For batches with $>70\%$ valid outputs, invalid outputs are iteratively replaced until batch validity reaches $>95\%$. Validation checks ensure that generated personas satisfy demographic and schema constraints, avoiding impossible combinations such as a child with a bachelor's degree and a high income level, following the validation approach described in \cite{marigliano2025aurora}. The hierarchy therefore helps keep generated responses consistent with the persona’s assigned demographic, ideological, and personality profile.

\section{Model: Event-Conditioned Emotional Response for Synthetic Personas}
Using the synthesized personas, we model each persona's event-conditioned response to a particular news event.

\paragraph{Event-Conditioned Response}
The final affective state of a persona is the outcome of three interacting layers: personality traits $\mathbf{o}$, moral foundations $\mathbf{m}$, and event--conditioned stance $s_{\text{pers}}(t)$. Personality provides a baseline tendency, while ideology and event stance influence how the persona responds emotionally to the event. Together, these yield the emotion vector
\[
\mathbf{e}(t) = f_{\text{emo}}\!\big(\mathbf{o}, \mathbf{m}, s_{\text{pers}}(t), u(t)\big),
\]
which represents the persona’s event-conditioned response to a particular news event.

\paragraph{Stage 1: Trait-Level Baseline}
The OCEAN personality vector $\mathbf{o} = [O,C,E,A,N]^\top$ is projected into Plutchik's eight emotion dimensions:
\begin{equation}
\mathbf{e}_{\text{base}} = \mathbf{W}\mathbf{o},
\end{equation}
where $\mathbf{W} \in \mathbb{R}^{8 \times 5}$ encodes directional trait--affect relationships grounded in personality psychology. This baseline represents stable predispositions rather than final expressed emotions. The mappings follow prior work linking Extraversion to positive affect and Joy \cite{watson1988panas,mccrae1992introduction,mcneil2006extraversion}, Neuroticism to negative affect, Fear, Sadness, and Anger \cite{watson1988panas,costa1992normal,mccrae1992introduction}, Agreeableness to cooperation and Trust \cite{mccrae1992introduction,costa1992normal}, Conscientiousness to planning and Anticipation \cite{costa1992normal,gross2001emotion}, and Openness to novelty and Surprise \cite{mccrae1992introduction,costa1992normal}. The weights are bounded calibration coefficients, not direct empirical estimates: stronger values are used for robust links such as Neuroticism to negative affect ($w=0.6$), moderate values for primary positive, prosocial, and novelty mappings ($w=0.4$), and weaker values for more indirect regulatory effects ($w=0.3$). Secondary cross-loadings, such as Neuroticism reducing Joy ($w=-0.3$), capture additional directional effects. Values are bounded to $[0,1]$.

\paragraph{Stage 2: Event-Conditioned Emotion Modulation}
The baseline is modulated by situational factors to produce context-dependent responses. Topic salience $u(t) \in [0,1]$ and opinion stance $s(t) \in [-1,1]$ jointly modulate affect with scaling strength $\alpha = 0.25$. Positive-valence emotions increase with supportive stances:
\begin{equation}
e_{\text{pos}}(t) = e_{\text{pos}}^{\text{base}} + \alpha \cdot u(t) \cdot s(t), \quad \text{pos} \in \{\text{joy, trust, anticipation}\},
\end{equation}
while negative-valence emotions are inversely modulated:
\begin{equation}
e_{\text{neg}}(t) = e_{\text{neg}}^{\text{base}} - \alpha \cdot u(t) \cdot s(t), \quad \text{neg} \in \{\text{anger, disgust, sadness, fear}\}.
\end{equation}
Surprise attenuates with salience independently of stance:
\begin{equation}
e_{\text{surprise}}(t) = e_{\text{surprise}}^{\text{base}} - \alpha \cdot (2u(t) - 1).
\end{equation}
All values remain in $[0,1]$ after modulation. Supportive stances ($s > 0$) amplify positive emotions and dampen negative ones; oppositional stances ($s < 0$) reverse this pattern. This modulation structure allows the same persona to produce different emotional responses across events while preserving stable trait-level tendencies. The goal is not to model human cognition directly, but to introduce interpretable event sensitivity into the generated responses.

\subsection{Opinion Formation}
Event-specific stances are derived as a convex combination of country and community priors, with individual variance:
\begin{multline}
s_{\text{pers}}(t) = \lambda_{\text{comm}} s_{\text{comm}}(t) + \lambda_{\text{ctry}} s_{\text{ctry}}(t) + \varepsilon,\\
\lambda_{\text{comm}} + \lambda_{\text{ctry}} \leq 1.
\end{multline}

Here, $s_{\text{pers}}(t) \in [-1,1]$ is persona $p$'s event-specific stance toward topic $t$, where negative values indicate opposition and positive values indicate support. The terms $s_{\text{comm}}(t)$ and $s_{\text{ctry}}(t)$ are the corresponding community-level and country-level stance priors for the same topic. The weights $\lambda_{\text{comm}}$ and $\lambda_{\text{ctry}}$ determine the relative influence of community and national context, while $\varepsilon \sim \mathcal{N}(0,\sigma_s^2)$ captures individual-level variation with variance $\sigma_s^2$.

\section{Experimental Design}
\label{sec:experiments}

\begin{figure}[htbp]
\centering
\resizebox{0.55\columnwidth}{!}{%
\begin{tikzpicture}[
    node distance=3mm,
    >=Latex,
    font=\scriptsize,
    proc/.style={
        rectangle,
        rounded corners,
        draw,
        align=center,
        minimum height=2mm,
        minimum width=26mm,
        inner sep=1mm
    },
    line/.style={-Latex, thick}
]

\node[proc] (personas) {Input Personas\\(Canada \& US)};
\node[proc, below=of personas] (prompt) {Persona-Specific\\Prompt Construction};
\node[proc, below=of prompt] (llm) {LLM Response\\Generation};
\node[proc, below=of llm] (emotion) {Sentiment \&\\Emotion Analysis};
\node[proc, below=of emotion] (interaction) {Community Interaction\\\& Context Update};
\node[proc, below=of interaction] (db) {Database Storage};

\draw[line] (personas) -- (prompt);
\draw[line] (prompt) -- (llm);
\draw[line] (llm) -- (emotion);
\draw[line] (emotion) -- (interaction);
\draw[line] (interaction) -- (db);

\draw[line, black] (db.west)
    .. controls +(-1.8,0) and +(-1.8,0)
    .. (prompt.west);

\node[align=center,
    font=\scriptsize\color{black},
    left=14mm of emotion
] {Next\\Timestep};

\end{tikzpicture}%
}
\caption{Experimental pipeline. Prompts combine persona attributes with event context. Responses propagate through the co-membership network, update neighboring personas at the next timestep, and are analyzed for emotion before the next cycle.}
\label{fig:experiment_pipeline}
    \Description{Block diagram of the experimental loop. Persona attributes
      and event context are combined into a prompt; the language model
      generates a response; the response propagates through the co-membership
      network to update neighbouring personas at the next timestep; emotion
      analysis of the responses feeds the next cycle.}
\end{figure}
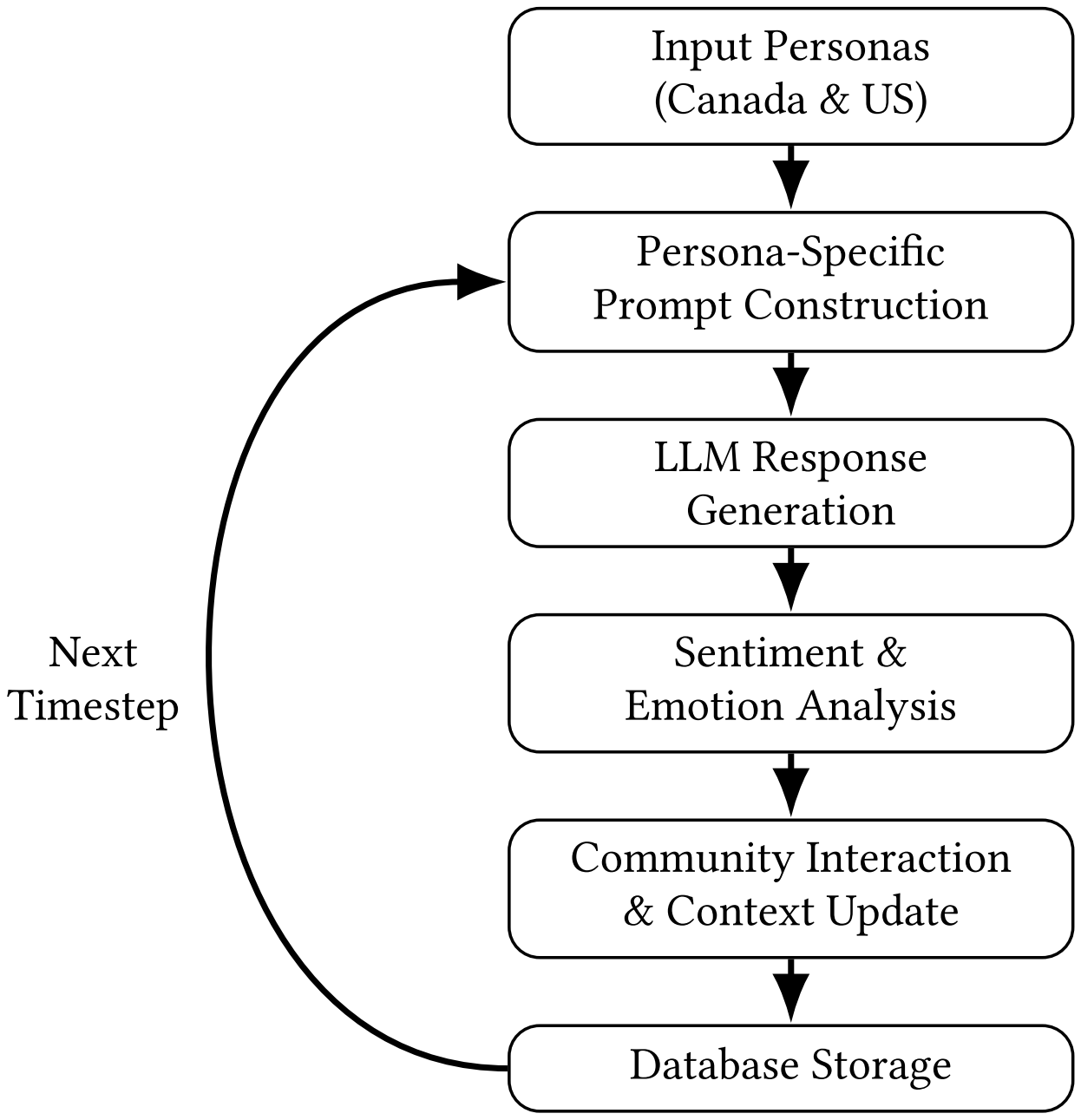
The experimental design evaluates how synthetic personas respond emotionally to event-conditioned prompts. Figure~\ref{fig:experiment_pipeline} summarizes the interaction cycle. At each timestep, persona-specific prompts are constructed, responses are generated, emotional content is analyzed, and the resulting posts are disseminated through the community network to update the conversational context for the next timestep.

\begin{enumerate}
    \item Construct persona-specific prompts using persona attributes, event information, and conversational context.
    \item Generate one response per persona using the shared LLM backend.
    \item Analyze the emotional content of each generated response and store the results.
    \item Disseminate the generated responses through the group co-membership network to update each persona's local conversational context for the next interaction timestep.
\end{enumerate}

\subsection{Prompt Construction}
\label{sec:prompt_construction}
Each prompt combines four information sources: (i) demographic attributes, (ii) psychological attributes determined by the experimental condition, (iii) event-specific information including stance and salience, and (iv) the persona's current conversational context. Experimental conditions ($\mathcal{L}$) differ only in the psychological attributes included (Table~\ref{tab:exp_levels}), while all other prompt components remain fixed. Complete prompt templates are provided in Appendix~\ref{appendix:prompts}.

 \subsection{Response Generation}

Each persona generates one social-media-style response per interaction timestep using the shared OpenAI GPT-4o-mini backend. Responses are conditioned on the complete persona specification, the current event, and the accumulated conversational context, while demographic and psychological attributes remain fixed throughout the simulation.

\subsection{Emotion Analysis}
Immediately after generation, responses are analyzed using a Plutchik-based emotion classifier. The resulting emotion scores are compared with each persona's baseline ($\mathbf{e}_{\text{base}}$) and event-conditioned ($\mathbf{e}(t)$) emotion profiles to evaluate persona consistency and support the validation measures described in Section~\ref{sec:quality_measures}.

\subsection{Social Interaction}
Generated posts are propagated through the group co-membership network and incorporated into the local conversational context of neighboring personas. This updated context serves as input for the subsequent interaction timestep, allowing discussions to evolve over ten rounds while demographic, ideological, and psychological attributes remain unchanged.

\section{Experimental Conditions}
Experiments are conducted using the personas described in Section~\ref{sec:methods}. Personas are assessed across five prompting levels ($\mathcal{L}$), two high-salience sociopolitical scenarios ($\mathcal{S}$), and $|T|=10$ interaction timesteps to examine how contextual information and event framing influence emotional responses over repeated interactions.

\subsection{Scenarios ($\mathcal{S}$).}
We consider two salient and ideologically distinct scenarios:
\begin{itemize}
\item \textbf{Roe v.~Wade Overturn (June 2022):} A polarized Supreme Court decision on reproductive rights, used to evaluate how political orientation and moral foundations shape emotional responses.
\item \textbf{988 Suicide Hotline (July 2022):} A mental health policy intervention with broad public relevance, used to examine responses to public health policy.
\end{itemize}

The Suicide Hotline scenario complements the primary scenario by representing an event with broad societal relevance and comparatively limited ideological polarization. The two scenarios also differ in both engagement patterns and demographic response distributions.

\subsection{Experiment Levels ($\mathcal{L}$).}
Table~\ref{tab:exp_levels} defines five persona prompting levels that progressively enrich the information provided to the LLM. Starting from a demographic-only baseline, successive levels add baseline emotions, Moral Foundations, OCEAN personality traits, and event-conditioned emotion cues, enabling a systematic ablation of prompt complexity. Prompting conditions are denoted by the fields $D$ (demographics), $B$ (baseline emotions), $M$ (Moral Foundations), $O$ (OCEAN personality traits), and $E$ (event-conditioned emotion cues).
\revised{The ablation-condition axes of the emotion heatmaps were generated with
an earlier label for the full condition; the axis entry \mbox{\texttt{D+M+O+E}} in
those panels denotes the \mbox{\texttt{D+M+O+E}} condition defined here.}

\begin{table}[htbp]
\centering
\caption{Persona prompting levels ($\mathcal{L}$) and fields provided to the model.}
\label{tab:exp_levels}
\footnotesize
\begin{tabular}{p{0.18\linewidth} p{0.45\linewidth} p{0.22\linewidth} p{0.15\linewidth}}
\toprule
\textbf{Condition} & \textbf{Description} & \textbf{Fields Included}  \\
\midrule
D & No emotional or psychological context. & -- \\
D+B & Adds baseline emotions from trait--affect mapping. & Baseline emotions \\
D+B+M & Adds moral foundations to baseline emotions. & Baseline emotions; moral values \\
D+B+M+O & Adds OCEAN traits to baseline emotions and moral values. & Baseline emotions; moral values; personality traits \\
D+M+O+E & Combines moral values, event-conditioned emotion cues, and OCEAN traits. & Moral values; event-conditioned cues; personality traits \\
\bottomrule
\end{tabular}
\end{table}

\subsection{Virtual Experiment}
The factorial design yields 53,700 persona responses. The total number of observations is determined by the cross-product of personas $\mathcal{P}$, scenarios $\mathcal{S}$, experiment levels $\mathcal{L}$, and interaction timesteps $T$:
\begin{equation}
N_{\text{exp}} = |\mathcal{P}| \times |\mathcal{S}| \times |\mathcal{L}| \times |T|.
\end{equation}

The repeated-response design over $n=10$ timesteps enabled evaluation of behavioral consistency across evolving conversational contexts. 

Although personas ($n=537$, see \autoref{tab:persona_summary}) remain in the same communities throughout the simulation, the evolving conversational context makes responses generated at different timesteps distinct experimental observations.

\section{Validation}
\label{sec:quality_measures}
We adopted a two pronged approach to validation. For determining internal validity, we compute quality measures over generated responses \cite{marigliano2025aurora} (see Appendix \ref{appendix:internal_consistency}). For external validity, we compare the rates of emotional expression in generated responses with emotional engagement data drawn from social media.

\subsection{Response Quality}
Response quality was evaluated post-hoc using automated 0--1 metrics computed from stored responses, without human annotation or LLM-as-judge evaluation. Emotional intensity was the mean normalized score across Plutchik's eight emotions extracted by a transformer-based emotion classifier:\[
I_{\text{emo}} = |\mathcal{E}|^{-1}\sum_{e \in \mathcal{E}} s_e.
\]

Topical relevance measured lexical overlap between response and scenario keywords:
\[
R_{\text{topic}} = \frac{|\mathcal{K}_r \cap \mathcal{K}_s|}{|\mathcal{K}_s|}.
\]

Here, $\mathcal{K}_r$ denotes the set of keywords extracted from the generated response, and $\mathcal{K}_s$ denotes the set of scenario-specific keywords used to represent the event. Keywords are extracted using TF-IDF scoring with stopword removal, selecting the top 10 most distinctive terms from each response and scenario description.

Persona consistency measured the extent to which the generated emotional response remained consistent with the persona's assigned psychological profile. Specifically, it compared the expressed emotion distribution with the persona's expected baseline emotional profile derived from the OCEAN personality traits: 

\[
PC_{\text{cons}} =
1 - |\mathcal{E}|^{-1}
\sum_{e \in \mathcal{E}}
|s_e-\hat{s}_e|,
\]

where $\hat{s}_e$ denotes the persona's baseline emotion score. This metric evaluates internal consistency between the assigned persona attributes and the generated response rather than agreement with human emotional responses.

Overall quality was computed as the mean of the three metrics:\[
Q = \frac{I_{\text{emo}} + R_{\text{topic}} + PC_{\text{cons}}}{3}.
\] Because these measures used rule-based lexical analysis and a separate fine-tuned emotion classifier rather than the generative LLM, they reduce the risk of circular validation.

\subsection{Validation Against Social Media Data}

Finally, we compare the distribution of emotional expression in LLM responses with Facebook engagement data for each event. Facebook reactions are treated as observable indicators of affective engagement rather than direct measures of emotional state. Because reactions (e.g., Love, Angry, Wow, and Sorry) do not map directly to Plutchik emotion categories, comparisons are made at the demographic and population level rather than through one-to-one emotion matching. Using the Social Science One URL Shares dataset, which provides aggregate counts of Facebook reactions \cite{socialscienceone2023urls}, we query event-related URLs, group reactions by age, gender, and political affiliation, and compare the resulting demographic distributions with those generated by the LLM personas. This evaluation therefore assesses whether the synthetic personas reproduce population-level patterns of emotional expression, rather than validating individual emotional states.

\section{Results}
\label{sec:results}
\subsection{Institutional Community Generation and Validation}

Generation outputs are summarized in Section~\ref{sec:methods} and Table~\ref{tab:persona_summary}. We produced $N{=}537$ personas (US: $267$, Canada: $270$) across six institutional categories, with balanced category representation (16.4--16.8\%) and consistent community sizes (8--9 personas). The sample closely matched the selected demographic reference distributions, while ideological and psychological attributes followed the specified MFT and OCEAN priors.

To further verify that the generated personas remain internally consistent with the intended hierarchical generation process, we performed additional validation of demographic fidelity, community-level consistency, and opinion generation in Appendix~\ref{appendix:internal_consistency}. They also extend the internal validation framework already conducted in AURORA \cite{marigliano2025aurora}, which demonstrated demographic plausibility, psychological coherence, and salience-aware opinion generation for synthetic populations.

\begin{figure}[htbp]
\centering
\includegraphics[width=0.95\linewidth]{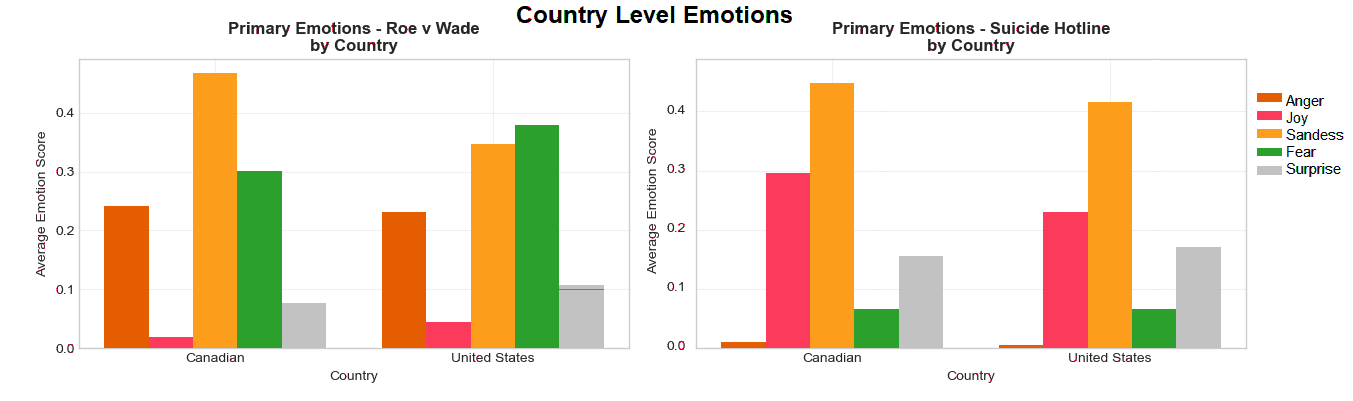}
\caption{Country-level primary emotions for Roe v.~Wade (left) and Suicide Hotline 988 (right).}
\label{fig:country_emotions}
    \Description{Two panels of grouped bar charts showing country-level
      primary emotion frequencies for United States and Canadian personas, for
      Roe v. Wade on the left and Suicide Hotline 988 on the right.}
\end{figure}

\subsection{Country-Level Emotions}

Figure~\ref{fig:country_emotions} shows mean emotion scores by country. For Roe v.~Wade, Canadian personas show higher sadness than U.S.\ personas (0.467 vs.\ 0.347), while U.S.\ personas show higher fear (0.379 vs.\ 0.300). Anger is nearly identical across countries (0.244 vs.\ 0.238). Thus, Canadian responses are more sadness-weighted, while U.S.\ responses show comparatively higher fear.

For the Suicide Hotline 988 stimulus, sadness is dominant in both countries. Canadian personas show slightly higher sadness (0.445 vs.\ 0.421) and joy (0.296 vs.\ 0.229), while anger remains negligible in both populations. Overall, Roe v.~Wade produces a more polarized sadness/fear profile, whereas the Suicide Hotline event produces sadness-oriented and prosocial responses.

\begin{figure}[htbp]
\centering
\includegraphics[width=0.85\linewidth]{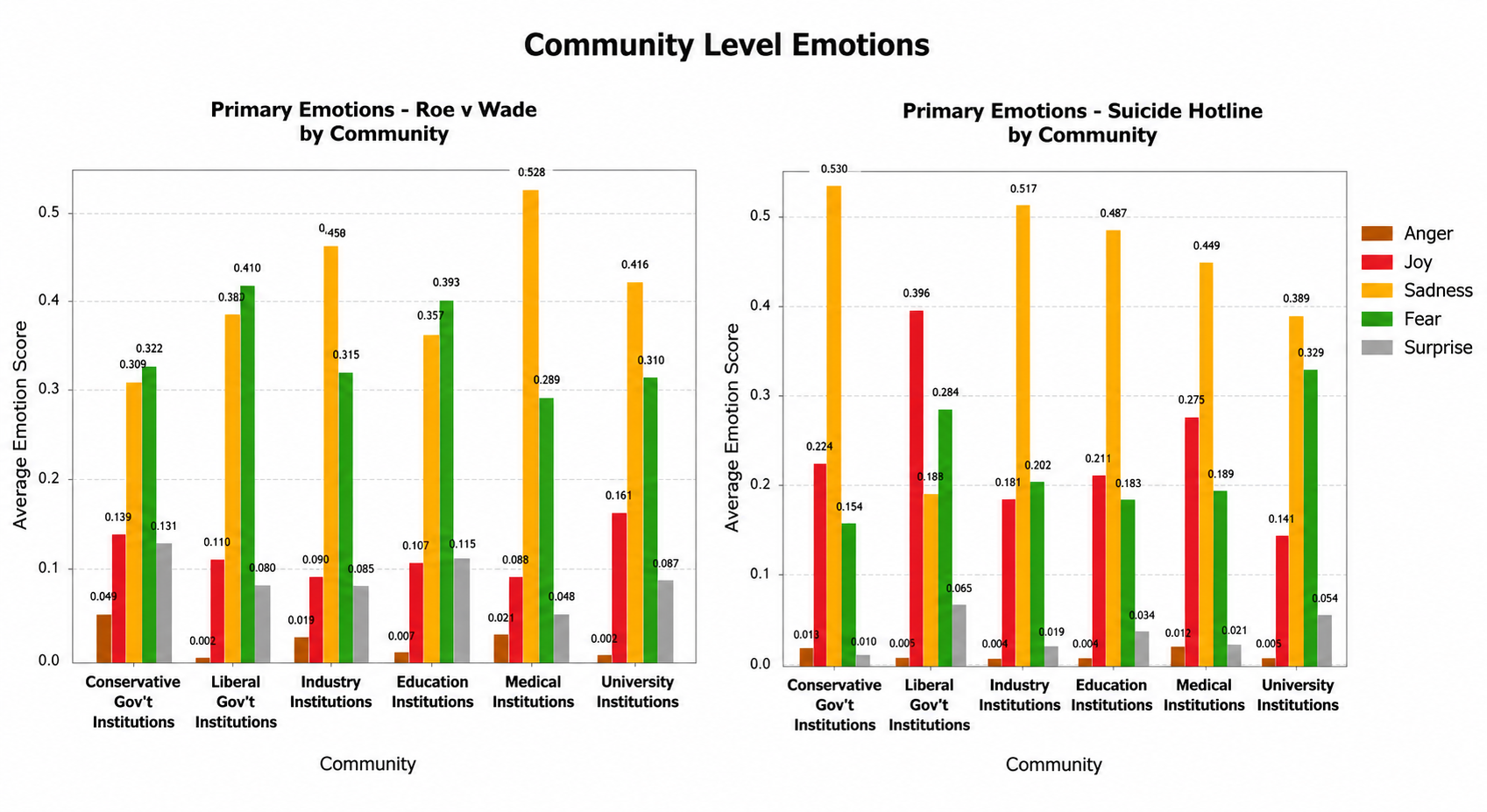}
\caption{Community-level primary emotions for Roe v.~Wade (left) and Suicide Hotline 988 (right).}
\label{fig:community_emotions}
    \Description{Two panels of grouped bar charts showing primary emotion
      frequencies broken down by institutional community type, for Roe v. Wade
      on the left and Suicide Hotline 988 on the right.}
\end{figure}

\subsection{Community-Level Emotions}

Figure~\ref{fig:community_emotions} reports mean emotion scores by institutional community. 
For Roe v.~Wade, sadness is the strongest response across most communities, especially medical (\revised{0.528}), industry (0.458), and university institutions (0.416). Fear is also elevated, particularly in liberal governmental (\revised{0.410}), education (0.393), and conservative governmental institutions (0.322). \revised{Anger remains low overall, with the highest values in conservative governmental (0.049), medical (0.021), and industry institutions (0.019).}

For the Suicide Hotline 988 stimulus, anger remains low across communities. \revised{Sadness is highest in conservative governmental (0.530), industry (0.517), education (0.487), and medical institutions (0.449), consistent with the country-level result that sadness dominates this event. Fear is markedly lower, peaking in university (0.329) and liberal governmental institutions (0.284), while joy is comparatively stronger in liberal governmental (0.396), medical (0.275), and conservative governmental institutions (0.224).} Community-level responses differed substantially across the two events.

\subsection{Individual-Level Emotions}

At the individual level, Table~\ref{tab:individual_emotions} reports representative persona emotion vectors. For Roe v.~Wade, Ethan Roberts (U.S., Far Left) shows a strongly negative profile, with high anger (0.629), sadness (0.584), fear (0.578), surprise (0.591), and disgust (0.510). This is consistent with his oppositional stance ($s=-0.965$) and high salience ($u=0.81$). In contrast, Amina N'Diaye (Canada, Center Left) shows a more moderate sadness-weighted response, with sadness (0.460), fear (0.390), and joy (0.360).

\begin{table}[htbp]
\centering
\caption{Representative individual-level emotion scores by nationality and event.}
\label{tab:individual_emotions}
\scriptsize
\setlength{\tabcolsep}{3.5pt}
\begin{tabularx}{\columnwidth}{lcccccccc}
\toprule
\textbf{Persona (Country)} & \textbf{Joy} & \textbf{Trust} & \textbf{Fear} & \textbf{Sad.} & \textbf{Ang.} & \textbf{Ant.} & \textbf{Disg.} & \textbf{Surp.} \\
\midrule
\multicolumn{9}{l}{\textbf{Roe v.~Wade Overturn (June 2022)}} \\
Ethan Roberts (U.S.)       & 0.166 & 0.187 & 0.578 & 0.584 & 0.629 & 0.515 & 0.510 & 0.591 \\
Amina N'Diaye (Canada)     & 0.360 & 0.310 & 0.390 & 0.460 & 0.280 & 0.290 & 0.300 & 0.330 \\
\midrule
\multicolumn{9}{l}{\textbf{988 Suicide Hotline (July 2022)}} \\
Ethan Roberts (U.S.)       & 0.157 & 0.359 & 0.118 & 0.422 & 0.191 & 0.437 & 0.116 & 0.178 \\
Amina N'Diaye (Canada)     & 0.350 & 0.580 & 0.260 & 0.620 & 0.300 & 0.490 & 0.270 & 0.380 \\
\bottomrule
\end{tabularx}
\begin{tablenotes}
\scriptsize
\item Scores represent normalized emotion dimensions from representative individual persona analyses.
\end{tablenotes}
\end{table}

For the 988 Suicide Hotline stimulus, both personas shift toward a less polarized response. Ethan shows higher anticipation (0.437), sadness (0.422), and trust (0.359), while Amina shows high sadness (0.620), trust (0.580), anticipation (0.490), and joy (0.350). Overall, the individual examples align with the aggregate results: Roe v.~Wade elicits more polarized negative affect, while Suicide Hotline 988 elicits sadness, trust, and anticipatory support with lower anger.

\subsection{Hierarchical Prompt Engineering Validation: Content Quality and Event-Conditioned Emotion Cue Effectiveness}

This section validates the effectiveness of hierarchical prompt enrichment on LLM content quality and emotional persona consistency . A controlled experiment systematically compared five prompt complexity levels across politically salient topics.

\subsubsection{Experimental Framework and Quality Metrics}

The study followed an ablation design in which contextual features were added stepwise to assess their impact on content quality: starting with demographics only (D), then adding baseline emotions (D+B), moral foundations (D+B+M), empirically grounded OCEAN traits (D+B+M+O), and finally event-conditioned emotion cues (D+M+O+E). \footnote{The final condition omits baseline emotions (B) because event-conditioned emotion cues supersede them; including both was redundant and sometimes inconsistent in pilot prompts.}

\begin{table}[htbp]
\centering
\caption{Content performance across ablation conditions}
\label{tab:quality_progression}
\scriptsize
\begin{tabularx}{\linewidth}{lXXXX}
\toprule
\textbf{Condition} & \textbf{Emotional Intensity} & \textbf{Topical Relevance} & \textbf{Persona consistency Score} & \textbf{Overall Quality}$^a$ \\
\midrule
D & 0.086 & 0.414 & 0.363 & 0.287 \\
D+B & 0.172 & 0.488 & 0.424 & 0.361 \\
D+B+M & 0.236 & 0.516 & 0.485 & 0.412 \\
D+B+M+O & 0.253 & 0.534 & 0.514 & 0.434 \\
D+M+O+E & 0.288 & 0.577 & 0.580 & 0.482 \\
\midrule
\textbf{Improvement}$^b$ & \textbf{+237\%} & \textbf{+40\%} & \textbf{+60\%} & \textbf{+68\%} \\
\bottomrule
\end{tabularx}
\begin{tablenotes}
\footnotesize
\item[$^a$] Overall quality represents the mean across all metrics (0–1 scale).
\item[$^b$] Percentage improvement calculated relative to the baseline (D).
\end{tablenotes}
\end{table}

Content quality was evaluated on four 0--1 metrics---\textit{emotional intensity}, \textit{topical relevance}, \textit{persona consistency score}, and \textit{overall quality} (the mean of the first three), across 53,700 responses (10 per persona per condition). Table~\ref{tab:quality_progression} reports mean scores by ablation condition and shows monotonic improvements as contextual fields are added: relative to the demographic only baseline (D), emotional intensity increases by +237\%, persona consistency by +60\%, topical relevance by +40\%, and overall quality by +68\%, with the highest scores under event-conditioned emotion cues (D+M+O+E). 

These differences are statistically reliable (ANOVA: intensity $F=464.88$, relevance $F=139.81$, persona consistency $F=303.20$; all $p<0.001$), and Bonferroni-corrected pairwise tests consistently favor D+M+O+E over D with large effects ($d=1.72$ intensity; $d=1.37$ persona consistency; $d=0.97$ relevance), indicating both statistical and practical gains from prompt enrichment.

\begin{table}[htbp]
\centering
\caption{D+M+O+E event-conditioned emotional response advantages relative to baseline (D).}
\label{tab:advanced_advantages}
\footnotesize
\begin{tabular}{p{0.2\linewidth}p{0.20\linewidth}rr}
\toprule
\textbf{Topic} & \textbf{Emotion} & \textbf{D+M+O+E Score} & \textbf{Change vs D} \\
\midrule
\multirow{3}{*}{Roe v.~Wade} & Fear & 0.457 & +665\% \\
 & Sadness & 0.566 & +464\% \\
 & Anger & 0.100 & --56\% \\
\midrule
\multirow{3}{*}{Suicide Hotline} & Joy & 0.441 & +528\% \\
 & Fear & 0.152 & +271\% \\
 & Neutral & 0.022 & --97\% \\
\bottomrule
\end{tabular}
\end{table}
\subsubsection{Event-Conditioned Emotional Response Enhancement and Distributional Shifts}

Event-conditioned emotion cue prompting (D+M+O+E) produced the clearest gains in event-aligned affect relative to demographic only prompting (D) (Table~\ref{tab:advanced_advantages}). For \textbf{Roe v.~Wade}, the enriched prompts shift responses toward threat- and loss-oriented emotions, increasing fear (0.060 $\rightarrow$ 0.457; +665\%) and sadness (0.100 $\rightarrow$ 0.566; +464\%), while reducing anger (0.227 $\rightarrow$ 0.100; --56\%). For \textbf{988 Suicide Hotline}, D+M+O+E increases engagement and prosocial tone, boosting joy (0.070 $\rightarrow$ 0.441; +528\%) and fear (0.041 $\rightarrow$ 0.152; +271\%), and sharply reducing neutral responses (0.675 $\rightarrow$ 0.022; --97\%). Together, these shifts indicate that event-conditioned emotion cues steer model outputs toward more context-appropriate affect rather than uniformly increasing intensity.

\begin{figure}[htbp]
    \centering
    \includegraphics[width=0.75\linewidth]{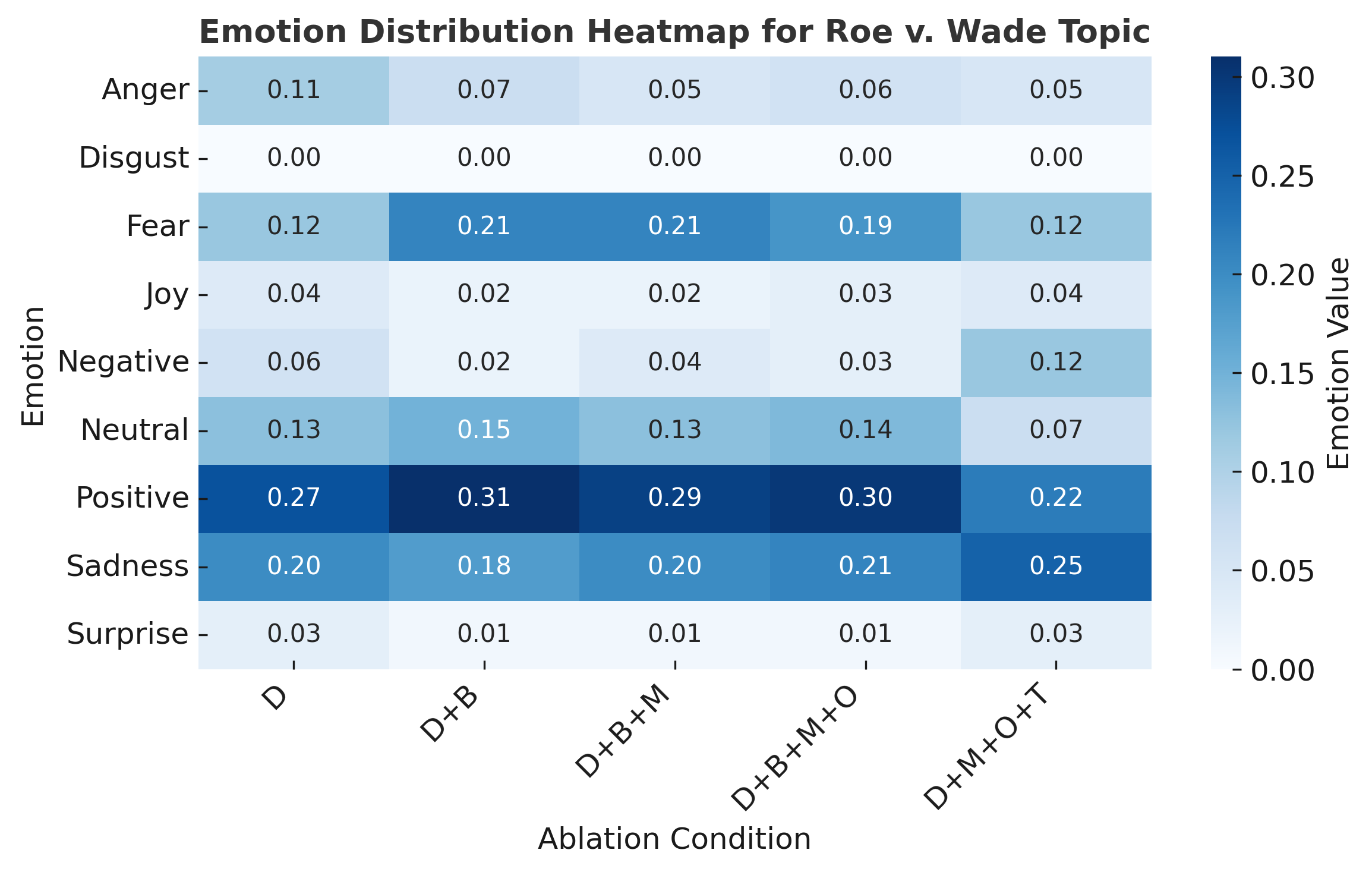}
    \caption{Emotion distribution heatmap for Roe v.~Wade topic by ablation condition.}
    \label{fig:emotion_heatmap_roe}
    \Description{Heatmap of the emotion distribution for the Roe v. Wade
      topic. Rows are ablation conditions and columns are Plutchik emotions;
      cell shading encodes the proportion of responses expressing each
      emotion.}
\end{figure}

\begin{figure}[htbp]
    \centering
    \includegraphics[width=0.75\linewidth]{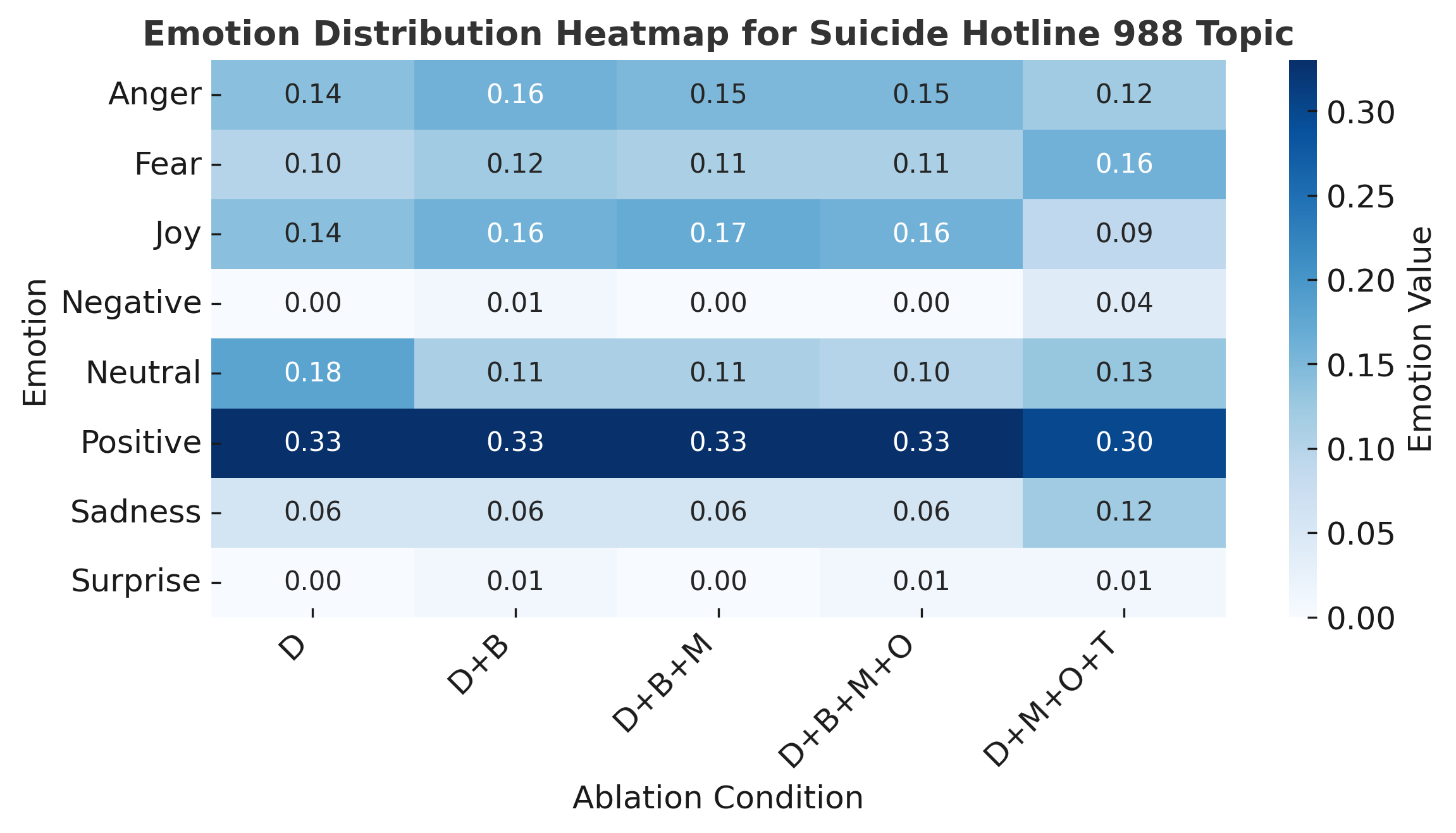}
    \caption{Emotion distribution heatmap for Suicide Hotline 988 topic by ablation condition.}
    \label{fig:emotion_heatmap_988}
    \Description{Heatmap of the emotion distribution for the Suicide Hotline
      988 topic. Rows are ablation conditions and columns are Plutchik
      emotions; cell shading encodes the proportion of responses expressing
      each emotion.}
\end{figure}

Figures~\ref{fig:emotion_heatmap_roe}--\ref{fig:emotion_heatmap_988} show mean emotion proportions by ablation condition, revealing how the distribution shifts as additional persona fields are introduced. For \textit{Roe v.~Wade}, sadness increases across conditions (0.20 $\rightarrow$ 0.25), fear remains in a moderate band (0.12--0.21), and anger declines (0.11 $\rightarrow$ 0.05); neutral responses also drop (0.13 $\rightarrow$ 0.07), indicating fewer low-affect outputs. For \textit{988 Suicide Hotline}, neutral/disengaged responses decrease (0.18 $\rightarrow$ 0.13) while sadness (0.06 $\rightarrow$ 0.12) and fear (0.10 $\rightarrow$ 0.16) increase modestly. Joy rises at intermediate conditions (peak 0.17) and then decreases under D+M+O+E (0.09), yielding a less uniformly positive distribution and a more mixed affect profile for the topic.

Statistical testing reinforced these findings: across 140 pairwise comparisons, 36 (25.7\%) reached significance after Bonferroni correction---well above chance. Effect sizes ranged from small to large, with the strongest effects concentrated in emotional intensity and persona consistency metrics.  

These results should not be interpreted as proving that D+M+O+E produces the uniquely ``correct'' emotional response. Similarly, improvements in persona consistency indicate greater alignment with the assigned persona specification rather than closer correspondence to human emotional responses. Rather, the ablation results show that event-conditioned prompting improves response structure, reduces neutral or low-information outputs, and shifts emotion distributions toward event-diagnostic affect. The enriched condition produces responses that are more emotionally differentiated and more aligned with the event context than the demographic only baseline, while still requiring external validation against observed human behavior.

\begin{figure}[htbp]
    \centering
    \includegraphics[width=.9\textwidth]{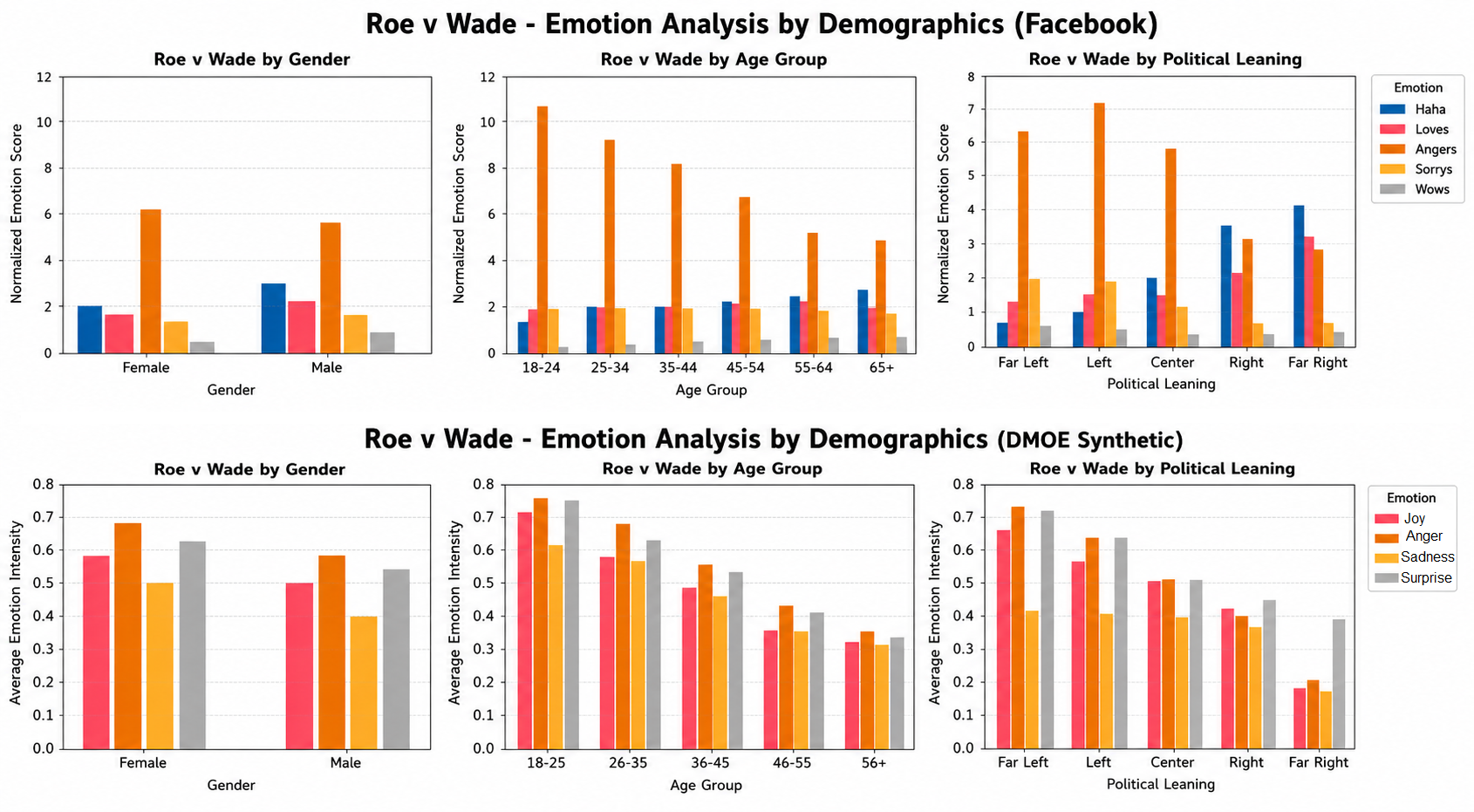}
    \caption{Facebook reaction patterns for Roe v.~Wade across demographic categories and (D+M+O+E) Synthetic persona event-conditioned response patterns for Roe v.~Wade across demographic categories}
    \label{fig:roe_v_wade_combined}
    \Description{Two rows of grouped bar charts for Roe v. Wade. The top row
      shows Facebook reaction shares broken down by gender, age group, and
      political leaning; the bottom row shows the corresponding synthetic
      persona emotional response shares under the D+M+O+E condition across the
      same three demographic breakdowns.}
\end{figure}





\subsection{Event-Conditioned Emotional Distributions}

\begin{figure}[htbp]
    \centering
    \includegraphics[width=.9\textwidth]{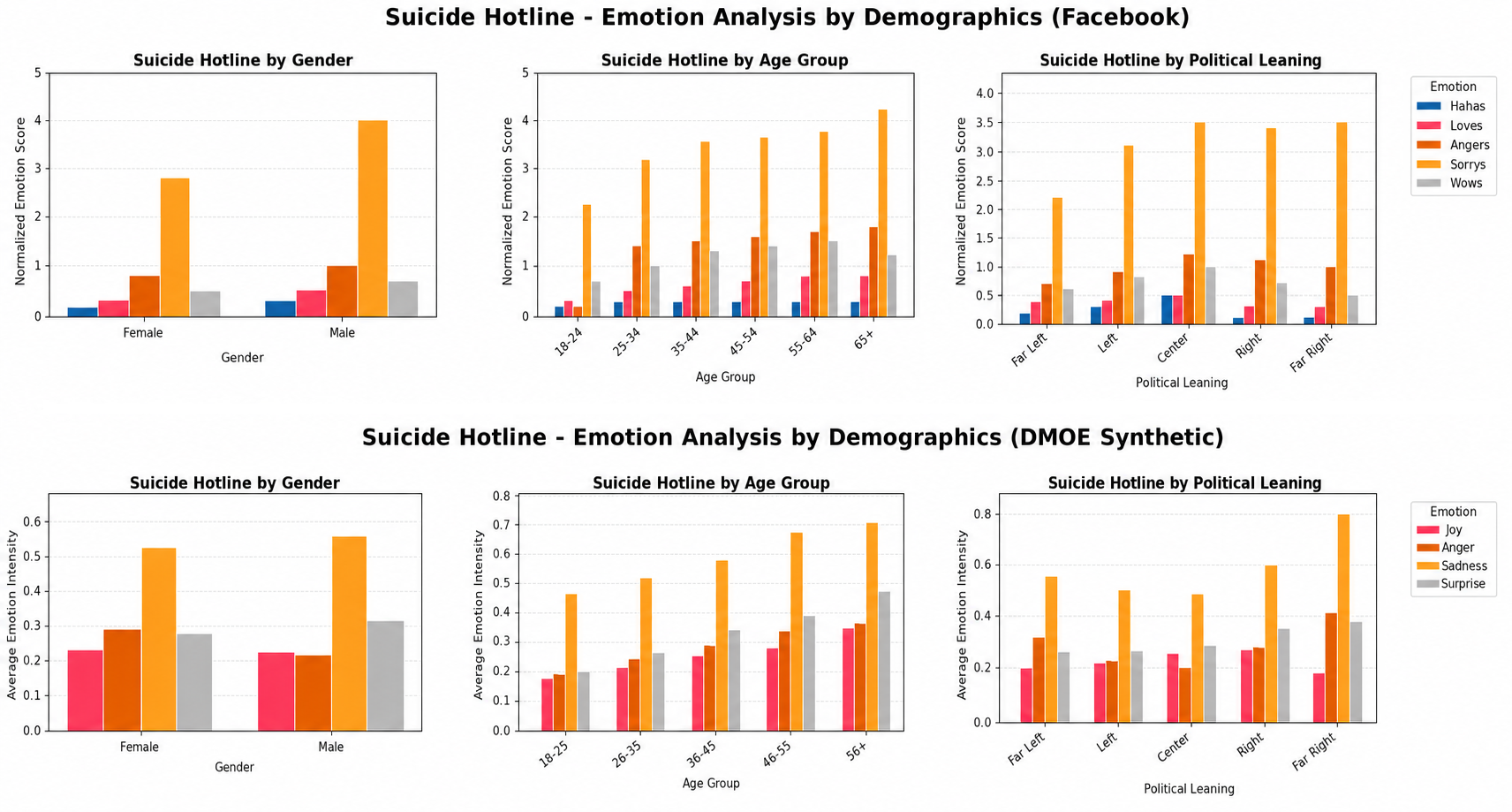}
    \caption{Facebook reaction patterns for Suicide Hotline 988 across demographic categories and (D+M+O+E) Synthetic persona event-conditioned response patterns for Suicide Hotline 988 across demographic categories}
    \label{fig:suicide_combined}
    \Description{Two rows of grouped bar charts for Suicide Hotline 988. The
      top row shows Facebook reaction shares broken down by gender, age group,
      and political leaning; the bottom row shows the corresponding synthetic
      persona emotional response shares under the D+M+O+E condition across the
      same three demographic breakdowns.}
\end{figure}

Figures~\ref{fig:roe_v_wade_combined}--\ref{fig:suicide_combined} show that event-conditioned responses vary by scenario and demographic group. For \textit{Roe v.~Wade}, anger decreases with age ($\beta_1=-0.847$, $p<0.001$), and political orientation separates responses, with stronger anger-related responses among left-leaning cohorts. For \textit{988 Suicide Hotline}, the dominant pattern shifts from anger to sadness: sadness increases with age, and political differences are smaller, reflecting a more empathetic and less polarized response. The Facebook and synthetic results also differ in distributional form. Facebook reactions are highly concentrated around the ``Sorry'' reaction, while LLM-generated personas spread expression across more emotions, though sadness remains dominant. The political-leaning plots show a related difference. In Facebook, left-leaning audiences show more anger and fewer ``Haha'' reactions, while right-leaning audiences show more ``Haha'' and less anger. This apparent tradeoff may partly reflect normalization, where increases in one reaction category reduce the relative share of others. In contrast, right-leaning synthetic personas mainly show lower overall emotional intensity rather than a clear anger-to-humor substitution. These results suggest that the prompting pipeline preserves interpretable demographic structure. Rather than producing a single generic emotional style, it generates event-conditioned responses that vary with age, ideology, and scenario context.

Additional validation using COVID-19 vaccine discourse and demographic emotional response comparisons is provided in Appendix~\ref{app:covid_validation}, where the synthetic personas similarly reproduce age-, gender-, and ideology-conditioned emotional trends observed in Facebook reaction data. Unlike the Roe v.~Wade and Suicide Hotline cases, the COVID-19 validation examines prolonged, high-salience public-health discourse characterized by both institutional trust dynamics and sustained political polarization.

\paragraph{Connection to Real-World Conflict Discourse}
The event sensitivity observed in the synthetic personas is consistent with prior findings from real conflict discourse. In bot-driven Twitter narratives during the Russia--Ukraine war, pro-Ukraine and pro-Russian messages differed not only by stance, but also by moral and affective framing: pro-Ukraine narratives emphasized solidarity and condemnation of Russian actions, while pro-Russian narratives framed intervention as protective and legitimate while degrading Ukraine and NATO \cite{marigliano_digitalpropaganda_2025}. Similarly, the personas produce event-conditioned emotional distributions, with more polarized anger- and fear-related responses for Roe v.~Wade and more sadness- and empathy-oriented responses for Suicide Hotline 988, rather than a single uniform emotional style.

\section{Discussion}
\label{discussion}

The results of this study indicate that hierarchically constructed LLM personas can reproduce several empirically observed patterns in event-conditioned response while preserving the practical advantages of simulation-based research. Validation against Facebook reaction data, combined with systematic ablation testing across prompt-complexity levels, provides support for the framework's behavioral plausibility. At the same time, the remaining distributional differences between empirical reactions and persona-generated responses point to important calibration targets for future work.

Age-dependent emotional regulation also emerged naturally within the synthetic population. Across the evaluated scenarios, older personas generally expressed less anger in response to highly polarizing events while exhibiting stronger sadness-oriented responses to mental-health-related topics. These context-dependent differences suggest that the hierarchical conditioning framework captures demographic variation in emotional expression rather than producing a single, uniform response pattern across all personas.

While the framework reproduced expected demographic variation, some divergences remained. For moralized issues such as Roe v.~Wade, personas showed slightly amplified emotional polarization, suggesting that LLMs may overemphasize strong moral cues relative to observed reactions. Older and highly conservative personas also showed compressed variability, indicating that the framework captures mean-level emotional patterns more effectively than fine-grained intragroup heterogeneity.

The framework was also sensitive to differences in event context. Highly polarizing issues, such as Roe v.~Wade, produced stronger ideological separation and more emotionally polarized responses, whereas the Suicide Hotline 988 scenario generated more gradual demographic variation centered on sadness, empathy, and support. The supplementary COVID-19 validation (Appendix~\ref{app:covid_validation}) reinforces this finding by showing that the personas preserve broad age, gender, and political-orientation trends within a prolonged public-health discussion. Together, these results suggest that the framework adapts emotional responses to the characteristics of the event rather than applying a fixed emotional profile across scenarios.

These results suggest that hierarchically conditioned synthetic populations can reproduce several interpretable demographic and emotional response patterns relevant to controlled simulation studies. Issues such as over-polarization in LLM outputs and homogeneity in smaller communities should be treated as directions for future calibration through improved prompting, feedback mechanisms, or model adjustment.

\subsection{Hierarchical Prompt Engineering and Prompt Structure}
The ablation results show why hierarchical prompting is effective. Adding demographic, moral, psychological, and event-conditioned emotion cues improved response quality, increasing emotional intensity by \revised{237\%} and persona consistency by 60\% from baseline to the full D+M+O+E condition. These gains did not simply make responses more emotional; they produced selective, context-appropriate shifts. \revised{For the polarizing Roe v.~Wade stimulus, fear and sadness rose sharply (+665\% and +464\%) while anger fell from 0.227 to 0.100 (\mbox{$-$56\%}); for the mental-health-related Suicide Hotline 988 stimulus, joy rose by 528\% and neutral responses fell by 97\% (\mbox{Table~\ref{tab:advanced_advantages}}). Conditioning therefore redirects affect toward event-appropriate emotions rather than amplifying all of them, and the decline in anger under the full condition is consistent with the under-expression of anger reported in the Abstract.}

These findings highlight the value of hierarchy. Country-level demographics provide macro-level structure, communities add local variation, and individual personas incorporate psychological depth through OCEAN traits, moral foundations, and event-conditioned priors. Together, these layers constrain outputs across demographic, ideological, and emotional dimensions, reducing incoherent persona behavior.

At the population level, the framework also reproduced social structure. Institutional community generation produced balanced category distributions (16.4--16.8\%) while preserving within-group variation. This shows that hierarchical prompting can model individual, group, and societal patterns within a consistent simulation framework. Overall, the ablation and demographic results support hierarchical prompt engineering as a useful method for building structured and empirically interpretable synthetic populations.

\subsection{Believability for Research Simulation}

In this work, realism refers to the preservation of interpretable demographic, ideological, and emotional response patterns rather than the exact reproduction of individual human behavior. The framework is intended to support controlled simulation studies by generating synthetic populations that preserve broad population-level patterns. Across the evaluated scenarios, the framework reproduced several observed demographic and topic-specific trends without manual parameter calibration. 

The framework produces demographic and emotional response patterns that are consistent with the evaluated scenarios, providing a practical approach. This method enables controlled social simulation without requiring human participants or sensitive personal data. By combining demographic, ideological, and psychological attributes, it enables the exploration of population-level responses to complex events. The hierarchical generation process can also be reweighted to represent census-aligned or underrepresented populations, allowing researchers to examine groups that are difficult to observe in online data.

\subsection{Scenarios and Interventions}

These results indicate that hierarchically constructed LLM personas can generate individual-level emotional responses and support the study of collective dynamics across scenarios. The framework couples macro demographic baselines with meso community structure and micro psychological traits, making the sources of variation explicit and the outputs more interpretable. The modular country$\rightarrow$community$\rightarrow$individual template also improves scalability by reusing shared context across personas under a fixed schema and validation layer.

These properties support scenario-based experimentation in which population responses to emerging events are simulated from empirically grounded community priors. The framework can therefore support controlled evaluation of crisis messaging, misinformation, and public-response scenarios, including counterfactual cases that would be difficult, costly, or unethical to study directly with human participants.

\section{Conclusions}

This study provides initial evidence that hierarchically conditioned LLM personas can reproduce several interpretable demographic and event-conditioned response patterns. By grounding personas in country-level context, community membership, psychological traits, and event-specific cues, the framework produced responses that aligned with several observed patterns in Facebook reaction data. Ablation testing showed that richer persona conditioning improved emotional intensity, topical relevance, and persona consistency, while the evaluated cases captured both polarizing dynamics on anger-prone topics (e.g., Roe v. Wade) and broader demographic gradients (e.g., increased sadness with age).

These findings suggest that hierarchical prompt engineering improves the behavioral plausibility of LLM-generated personas for synthetic social simulation. However, the current evaluation is limited to a small set of high-salience events in U.S. and Canadian contexts. Further validation across additional cultures, platforms, and event types is needed to assess the generalizability of the framework.

Overall, this work offers a foundation for extending deliberation-centric ABMs with agents capable of producing event-conditioned responses under structured prompting. Future work should expand validation across additional countries, platforms, time periods, and event types, and further calibrate emotion-specific parameters before applying the framework to high-stakes policy or intervention settings.

\subsection{Limitations}
Several limitations should be considered. First, the evaluation covers a limited set of events, so its generalizability to other cultural, political, and discourse contexts remains uncertain. Although the supplementary COVID-19 analysis broadens the scope, further longitudinal and cross-platform validation is needed. Second, LLM-generated personas may reflect biases from their training data and prompt design, affecting how demographic groups, moral values, and political orientations are represented \cite{bender2021dangers,weidinger2021ethical}. These concerns underscore the need for continued validation, calibration, and interpretability in synthetic social simulation. Third, the framework captures aggregate emotional and demographic patterns but does not yet model more complex dynamics, such as recursive reasoning, network contagion, or sustained temporal adaptation. The trait-to-affect projection also uses a linear weight matrix $\mathbf{W}$, although relationships between personality and affect may be non-linear. 

Critics caution that LLM-based personas may conflate linguistic coherence with psychological realism \cite{van_rooij_reclaiming_2024,schroder_large_2025}, lack the nuance needed for qualitative analysis \cite{kapania_simulacrum_2024}, or reflect implementation-specific artifacts rather than generalizable behavior \cite{ye2026stop}.
\revised{Accordingly}, all experiments used the same LLM backend (\revised{GPT-4o-mini}) to isolate the effects of hierarchical conditioning. For extensive cross-model evaluation, we refer the reader to original AURORA model, which forms the basis of this work \cite{marigliano2025aurora}.

At the same time, related work shows that structured and empirically grounded agent architectures can exhibit meaningful behavioral regularities \cite{park2023generative,argyle2023out,aher2023using,binz_centaur_2024,xie_can_2024}. Recent research also emphasizes the importance of validation and theoretical grounding in social-science applications \cite{davidson_integrating_2025}. Overall, the results support a middle ground: LLM personas should not replace human subjects, but they can help model collective behavior and test theory-driven hypotheses in controlled, ethical settings.

\subsection{Future Work}

Future research should validate the framework across more platforms, cultural contexts, and stimuli beyond politics and mental health. Longitudinal COVID-19 discourse offers a useful benchmark because it captures evolving narratives, shifting trust, and sustained crisis conditions. Future work should also add dynamic network processes, such as opinion diffusion and collective identity formation, and include bias auditing, fairness constraints, and real-world testing in crisis communication, misinformation response, and policy messaging. Because LLM outputs can diverge from empirical patterns, especially for negative emotions such as fear and anger, improved emotional calibration remains a key priority. These steps would strengthen the framework’s robustness, transparency, and ethical grounding.


\begin{acks}
The research for this paper was supported in part by the Knight Foundation, the Office of Naval Research under grant (N000142112229), the Army under grant (W911NF20D0002) through the AI2C center, and by the Center for Informed Democracy and Social-cybersecurity (IDeaS) and the Center for Computational Analysis of Social and Organizational Systems (CASOS) at Carnegie Mellon University. The views and conclusions are those of the authors. They should not be interpreted as representing the official policies, either expressed or implied, of the Knight Foundation, the Office of Naval Research, the US. Army, or the US Government.
\end{acks}

\section*{Data Availability Statement}

The Facebook URL Shares data were accessed through the Social Science One URL Shares Dataset via the Meta Researcher Platform under the applicable Research Data Agreement. Due to privacy and platform restrictions, raw data cannot be redistributed. Researchers may request access through Social Science One: \url{https://socialscience.one/request-proposals-facebook-urls-dataset}. This study reports only aggregated demographic-level summaries and no individual-level Facebook user data.

\bibliographystyle{ACM-Reference-Format}
\bibliography{corrected_references}

@misc{xie_can_2024,
	title = {Can {Large} {Language} {Model} {Agents} {Simulate} {Human} {Trust} {Behavior}?},
	url = {http://arxiv.org/abs/2402.04559},
	language = {en},
	urldate = {2024-11-04},
	publisher = {arXiv},
	author = {Xie, Chengxing and Chen, Canyu and Jia, Feiran and Ye, Ziyu and Lai, Shiyang and Shu, Kai and Gu, Jindong and Bibi, Adel and Hu, Ziniu and Jurgens, David and Evans, James and Torr, Philip and Ghanem, Bernard and Li, Guohao},
	month = nov,
	year = {2024},
	note = {arXiv:2402.04559 [cs]},
}

@misc{liu_improving_2023,
	title = {Improving {Interpersonal} {Communication} by {Simulating} {Audiences} with {Language} {Models}},
	url = {http://arxiv.org/abs/2311.00687},
	language = {en},
	urldate = {2024-11-04},
	publisher = {arXiv},
	author = {Liu, Ryan and Yen, Howard and Marjieh, Raja and Griffiths, Thomas L. and Krishna, Ranjay},
	month = nov,
	year = {2023},
	note = {arXiv:2311.00687 [cs]},
}

@misc{binz_centaur_2024,
	title = {Centaur: a foundation model of human cognition},
	shorttitle = {Centaur},
	url = {http://arxiv.org/abs/2410.20268},
	language = {en},
	urldate = {2024-11-04},
	publisher = {arXiv},
	author = {Binz, Marcel and Akata, Elif and Bethge, Matthias and Brändle, Franziska and Callaway, Fred and Coda-Forno, Julian and Dayan, Peter and Demircan, Can and Eckstein, Maria K. and Éltető, Noémi and Griffiths, Thomas L. and Haridi, Susanne and Jagadish, Akshay K. and Ji-An, Li and Kipnis, Alexander and Kumar, Sreejan and Ludwig, Tobias and Mathony, Marvin and Mattar, Marcelo and Modirshanechi, Alireza and Nath, Surabhi S. and Peterson, Joshua C. and Rmus, Milena and Russek, Evan M. and Saanum, Tankred and Scharfenberg, Natalia and Schubert, Johannes A. and Buschoff, Luca M. Schulze and Singhi, Nishad and Sui, Xin and Thalmann, Mirko and Theis, Fabian and Truong, Vuong and Udandarao, Vishaal and Voudouris, Konstantinos and Wilson, Robert and Witte, Kristin and Wu, Shuchen and Wulff, Dirk and Xiong, Huadong and Schulz, Eric},
	month = oct,
	year = {2024},
	note = {arXiv:2410.20268 [cs]},
}

@misc{kapania_simulacrum_2024,
	title = {'{Simulacrum} of {Stories}': {Examining} {Large} {Language} {Models} as {Qualitative} {Research} {Participants}},
	shorttitle = {'{Simulacrum} of {Stories}'},
	url = {http://arxiv.org/abs/2409.19430},
	doi = {10.48550/arXiv.2409.19430},
	language = {en},
	urldate = {2025-01-16},
	publisher = {arXiv},
	author = {Kapania, Shivani and Agnew, William and Eslami, Motahhare and Heidari, Hoda and Fox, Sarah},
	month = sep,
	year = {2024},
	note = {arXiv:2409.19430 [cs]},
}

@article{park_generative_nodate,
	title = {Generative {Agent} {Simulations} of 1,000 {People}},
	language = {en},
	author = {Park, Joon Sung and Zou, Carolyn Q and Shaw, Aaron and Hill, Benjamin Mako and Cai, Carrie and Morris, Meredith Ringel and Willer, Robb and Liang, Percy and Bernstein, Michael S},
}

@article{davidson_integrating_2025,
	title = {Integrating {Generative} {Artificial} {Intelligence} into {Social} {Science} {Research}: {Measurement}, {Prompting}, and {Simulation}},
	volume = {54},
	issn = {0049-1241},
	shorttitle = {Integrating {Generative} {Artificial} {Intelligence} into {Social} {Science} {Research}},
	url = {https://doi.org/10.1177/00491241251339184},
	doi = {10.1177/00491241251339184},
	language = {EN},
	number = {3},
	urldate = {2025-09-05},
	journal = {Sociological Methods \& Research},
	author = {Davidson, Thomas and Karell, Daniel},
	month = aug,
	year = {2025},
	note = {Publisher: SAGE Publications Inc},
	pages = {775--793},
}

@misc{schroder_large_2025,
	title = {Large {Language} {Models} {Do} {Not} {Simulate} {Human} {Psychology}},
	url = {http://arxiv.org/abs/2508.06950},
	doi = {10.48550/arXiv.2508.06950},
	language = {en},
	urldate = {2025-09-11},
	publisher = {arXiv},
	author = {Schröder, Sarah and Morgenroth, Thekla and Kuhl, Ulrike and Vaquet, Valerie and Paaßen, Benjamin},
	month = aug,
	year = {2025},
	note = {arXiv:2508.06950 [cs]},
}

@article{van_rooij_reclaiming_2024,
	title = {Reclaiming {AI} as a {Theoretical} {Tool} for {Cognitive} {Science}},
	volume = {7},
	issn = {2522-087X},
	url = {https://doi.org/10.1007/s42113-024-00217-5},
	doi = {10.1007/s42113-024-00217-5},
	language = {en},
	number = {4},
	urldate = {2025-09-11},
	journal = {Computational Brain \& Behavior},
	author = {van Rooij, Iris and Guest, Olivia and Adolfi, Federico and de Haan, Ronald and Kolokolova, Antonina and Rich, Patricia},
	month = dec,
	year = {2024},
	pages = {616--636},
}

@inproceedings{chopra2023agenttorch,
  title={Agenttorch: Agent-based modeling with automatic differentiation},
  author={Chopra, Ayush and Subramanian, Jayakumar and Krishnamurthy, Balaji and Raskar, Ramesh},
  booktitle={Second Agent Learning in Open-Endedness Workshop},
  year={2023}
}

@article{loewenstein2000emotions,
  title     = {Emotions in Economic Theory and Economic Behavior},
  author    = {Loewenstein, George},
  journal   = {American Economic Review},
  volume    = {90},
  number    = {2},
  pages     = {426--432},
  year      = {2000},
  doi       = {10.1257/aer.90.2.426},
  publisher = {American Economic Association}
}

@book{sapolsky2017behave,
  title     = {Behave: The Biology of Humans at Our Best and Worst},
  author    = {Sapolsky, Robert M.},
  year      = {2017},
  publisher = {Penguin Press},
  address   = {New York, NY}
}

@article{slovic2007affect,
  title     = {The Affect Heuristic},
  author    = {Slovic, Paul and Finucane, Melissa L. and Peters, Ellen and MacGregor, Donald G.},
  journal   = {European Journal of Operational Research},
  volume    = {177},
  number    = {3},
  pages     = {1333--1352},
  year      = {2007},
  doi       = {10.1016/j.ejor.2005.04.006},
  publisher = {Elsevier}
}

@article{czeisler2020mental,
  title   = {Mental Health, Substance Use, and Suicidal Ideation During the {COVID-19} Pandemic---United States, June 24--30, 2020},
  author  = {Czeisler, Mark {\'E}. and Lane, Rashon I. and Petrosky, Emiko and Wiley, Joshua F. and Christensen, Aleta and Njai, Rashid and Weaver, Matthew D. and Robbins, Rebecca and Facer-Childs, Elise R. and Barger, Laura K. and Czeisler, Charles A. and Howard, Mark E. and Rajaratnam, Shantha M. W.},
  journal = {Morbidity and Mortality Weekly Report},
  volume  = {69},
  number  = {32},
  pages   = {1049--1057},
  year    = {2020},
  doi     = {10.15585/mmwr.mm6932a1}
}

@incollection{plutchik1980general,
  title     = {A General Psychoevolutionary Theory of Emotion},
  author    = {Plutchik, Robert},
  booktitle = {Emotion: Theory, Research, and Experience. Volume 1: Theories of Emotion},
  editor    = {Plutchik, Robert and Kellerman, Henry},
  pages     = {3--33},
  year      = {1980},
  publisher = {Academic Press},
  address   = {New York, NY},
  isbn      = {978-0-12-558701-3}
}

@article{mohammad2013crowdsourcing,
  title   = {Crowdsourcing a Word--Emotion Association Lexicon},
  author  = {Mohammad, Saif M. and Turney, Peter D.},
  journal = {Computational Intelligence},
  volume  = {29},
  number  = {3},
  pages   = {436--465},
  year    = {2013},
  doi     = {10.1111/j.1467-8640.2012.00460.x}
}

@article{ekman1992argument,
  title     = {An Argument for Basic Emotions},
  author    = {Ekman, Paul},
  journal   = {Cognition \& Emotion},
  volume    = {6},
  number    = {3--4},
  pages     = {169--200},
  year      = {1992},
  doi       = {10.1080/02699939208411068},
  publisher = {Taylor \& Francis}
}

@article{jack2012facial,
  title     = {Facial Expressions of Emotion Are Not Culturally Universal},
  author    = {Jack, Rachael E. and Garrod, Oliver G. B. and Yu, Hui and Caldara, Roberto and Schyns, Philippe G.},
  journal   = {Proceedings of the National Academy of Sciences},
  volume    = {109},
  number    = {19},
  pages     = {7241--7244},
  year      = {2012},
  doi       = {10.1073/pnas.1200155109},
  publisher = {National Academy of Sciences}
}

@article{mohammad2016sentiment,
  title   = {Sentiment Analysis of Short Informal Texts},
  author  = {Kiritchenko, Svetlana and Zhu, Xiaodan and Mohammad, Saif M.},
  journal = {Journal of Artificial Intelligence Research},
  volume  = {50},
  pages   = {723--762},
  year    = {2014},
  doi     = {10.1613/jair.4272}
}

@article{graham2009liberals,
  title   = {Liberals and Conservatives Rely on Different Sets of Moral Foundations},
  author  = {Graham, Jesse and Haidt, Jonathan and Nosek, Brian A.},
  journal = {Journal of Personality and Social Psychology},
  volume  = {96},
  number  = {5},
  pages   = {1029--1046},
  year    = {2009},
  doi     = {10.1037/a0015141},
  publisher = {American Psychological Association}
}

@article{clifford2015moral,
  title     = {Moral Foundations Vignettes: A Standardized Stimulus Database of Scenarios Based on Moral Foundations Theory},
  author    = {Clifford, Scott and Iyengar, Vijeth and Cabeza, Roberto and Sinnott-Armstrong, Walter},
  journal   = {Behavior Research Methods},
  volume    = {47},
  number    = {4},
  pages     = {1178--1198},
  year      = {2015},
  doi       = {10.3758/s13428-014-0551-2},
  publisher = {Springer}
}

@article{van2014moral,
  title     = {The Moral Machine Experiment},
  author    = {Awad, Edmond and Dsouza, Sohan and Kim, Richard and Schulz, Jonathan and Henrich, Joseph and Shariff, Azim and Bonnefon, Jean-Fran{\c{c}}ois and Rahwan, Iyad},
  journal   = {Nature},
  volume    = {563},
  number    = {7729},
  pages     = {59--64},
  year      = {2018},
  doi       = {10.1038/s41586-018-0637-6},
  publisher = {Nature Publishing Group}
}

@article{mccrae1992introduction,
  title     = {An Introduction to the Five-Factor Model and Its Applications},
  author    = {McCrae, Robert R. and John, Oliver P.},
  journal   = {Journal of Personality},
  volume    = {60},
  number    = {2},
  pages     = {175--215},
  year      = {1992},
  doi       = {10.1111/j.1467-6494.1992.tb00970.x},
  publisher = {Wiley}
}

@article{costa1992normal,
  title   = {Normal Personality Assessment in Clinical Practice: The {NEO} Personality Inventory},
  author  = {Costa, Paul T. and McCrae, Robert R.},
  journal = {Psychological Assessment},
  volume  = {4},
  number  = {1},
  pages   = {5--13},
  year    = {1992},
  doi     = {10.1037/1040-3590.4.1.5}
}

@article{schmitt2007geographic,
  title     = {The Geographic Distribution of Big Five Personality Traits: Patterns and Profiles of Human Self-Description Across 56 Nations},
  author    = {Schmitt, David P. and Allik, J{\"u}ri and McCrae, Robert R. and Benet-Mart{\'i}nez, Ver{\'o}nica},
  journal   = {Journal of Cross-Cultural Psychology},
  volume    = {38},
  number    = {2},
  pages     = {173--212},
  year      = {2007},
  doi       = {10.1177/0022022106297299},
  publisher = {Sage Publications}
}

@book{watson2000mood,
  title     = {Mood and Temperament},
  author    = {Watson, David},
  publisher = {Guilford Press},
  address   = {New York, NY},
  year      = {2000}
}

@article{gross2001emotion,
  title     = {Emotion Regulation in Adulthood: Timing Is Everything},
  author    = {Gross, James J.},
  journal   = {Current Directions in Psychological Science},
  volume    = {10},
  number    = {6},
  pages     = {214--219},
  year      = {2001},
  doi       = {10.1111/1467-8721.00152},
  publisher = {Sage Publications}
}

@article{marigliano_digitalpropaganda_2025,
  author  = {Marigliano, Rebecca and Ng, Lynnette Hui Xian and Carley, Kathleen M.},
  title   = {Analyzing Digital Propaganda and Conflict Rhetoric: A Study on Russia's Bot-Driven Campaigns and Counter-Narratives During the Ukraine Crisis},
  journal = {Social Network Analysis and Mining},
  volume  = {14},
  number  = {170},
  year    = {2024},
  doi     = {10.1007/s13278-024-01322-w}
}

@inproceedings{marigliano2025aurora,
  author    = {Marigliano, Rebecca and Carley, Kathleen M.},
  title     = {{AURORA}: Enhancing Synthetic Population Realism Through {RAG} and Salience-Aware Opinion Modeling},
  booktitle = {Proceedings of the 2025 Winter Simulation Conference},
  editor    = {Azar, E. and Djanatliev, A. and Harper, A. and Kogler, C. and Ramamohan, V. and Anagnostou, A. and Taylor, S. J. E.},
  pages     = {2563--2574},
  year      = {2025},
  publisher = {IEEE},
  doi       = {10.1109/WSC68292.2025.11339080}
}

@article{mcneil2006extraversion,
  title   = {The Causal Effects of Extraversion on Positive Affect and Neuroticism on Negative Affect: Manipulating State Extraversion and State Neuroticism in an Experimental Approach},
  author  = {McNiel, J. Murray and Fleeson, William},
  journal = {Journal of Research in Personality},
  volume  = {40},
  number  = {5},
  pages   = {529--550},
  year    = {2006},
  doi     = {10.1016/j.jrp.2005.05.003}
}

@inproceedings{park2023generative,
  title     = {Generative Agents: Interactive Simulacra of Human Behavior},
  author    = {Park, Joon Sung and O'Brien, Joseph C. and Cai, Carrie J. and Morris, Meredith Ringel and Liang, Percy and Bernstein, Michael S.},
  booktitle = {Proceedings of the 36th Annual {ACM} Symposium on User Interface Software and Technology},
  series    = {{UIST} '23},
  articleno = {2},
  numpages  = {22},
  year      = {2023},
  publisher = {Association for Computing Machinery},
  address   = {New York, NY, USA},
  doi       = {10.1145/3586183.3606763}
}

@article{watson1988panas,
  title   = {Development and Validation of Brief Measures of Positive and Negative Affect: The {PANAS} Scales},
  author  = {Watson, David and Clark, Lee Anna and Tellegen, Auke},
  journal = {Journal of Personality and Social Psychology},
  volume  = {54},
  number  = {6},
  pages   = {1063--1070},
  year    = {1988},
  doi     = {10.1037/0022-3514.54.6.1063}
}

@article{argyle2023out,
  title     = {Out of One, Many: Using Language Models to Simulate Human Samples},
  author    = {Argyle, Lisa P. and Busby, Ethan C. and Fulda, Nancy and Gubler, Joshua R. and Rytting, Christopher and Wingate, David},
  journal   = {Political Analysis},
  volume    = {31},
  number    = {3},
  pages     = {337--351},
  year      = {2023},
  doi       = {10.1017/pan.2023.2},
  publisher = {Cambridge University Press}
}

@inproceedings{wei2022chain,
  title     = {Chain-of-Thought Prompting Elicits Reasoning in Large Language Models},
  author    = {Wei, Jason and Wang, Xuezhi and Schuurmans, Dale and Bosma, Maarten and Ichter, Brian and Xia, Fei and Chi, Ed H. and Le, Quoc V. and Zhou, Denny},
  booktitle = {Advances in Neural Information Processing Systems},
  volume    = {35},
  pages     = {24824--24837},
  year      = {2022},
  publisher = {Curran Associates, Inc.}
}

@inproceedings{zhou2022least,
  title     = {Least-to-Most Prompting Enables Complex Reasoning in Large Language Models},
  author    = {Zhou, Denny and Sch{\"a}rli, Nathanael and Hou, Le and Wei, Jason and Scales, Nathan and Wang, Xuezhi and Schuurmans, Dale and Cui, Claire and Bousquet, Olivier and Le, Quoc V. and Chi, Ed H.},
  booktitle = {The Eleventh International Conference on Learning Representations},
  year      = {2023},
  url       = {https://openreview.net/forum?id=WZH7099tgfM}
}

@article{bail2018exposure,
  title     = {Exposure to Opposing Views on Social Media Can Increase Political Polarization},
  author    = {Bail, Christopher A. and Argyle, Lisa P. and Brown, Taylor W. and Bumpus, John P. and Chen, Haohan and Hunzaker, M. B. Fallin and Lee, Jaemin and Mann, Marcus and Merhout, Friedolin and Volfovsky, Alexander},
  journal   = {Proceedings of the National Academy of Sciences},
  volume    = {115},
  number    = {37},
  pages     = {9216--9221},
  year      = {2018},
  doi       = {10.1073/pnas.1804840115},
  publisher = {National Academy of Sciences}
}

@article{brady2017mad,
  title     = {The {MAD} Model of Moral Contagion: The Role of Motivation, Attention, and Design in the Spread of Moralized Content Online},
  author    = {Brady, William J. and Crockett, M. J. and Van Bavel, Jay J.},
  journal   = {Perspectives on Psychological Science},
  volume    = {15},
  number    = {4},
  pages     = {978--1010},
  year      = {2020},
  doi       = {10.1177/1745691620917336},
  publisher = {Sage Publications}
}

@article{bazarova2012self,
  title     = {Self-Disclosure in Social Media: Extending the Functional Approach to Disclosure Motivations and Characteristics on Social Network Sites},
  author    = {Bazarova, Natalya N. and Choi, Yoon Hyung},
  journal   = {Journal of Communication},
  volume    = {64},
  number    = {4},
  pages     = {635--657},
  year      = {2014},
  doi       = {10.1111/jcom.12106},
  publisher = {Oxford University Press}
}

@article{kramer2014experimental,
  title     = {Experimental Evidence of Massive-Scale Emotional Contagion Through Social Networks},
  author    = {Kramer, Adam D. I. and Guillory, Jamie E. and Hancock, Jeffrey T.},
  journal   = {Proceedings of the National Academy of Sciences},
  volume    = {111},
  number    = {24},
  pages     = {8788--8790},
  year      = {2014},
  doi       = {10.1073/pnas.1320040111},
  publisher = {National Academy of Sciences}
}

@book{hofstede2001culture,
  title     = {Culture's Consequences: Comparing Values, Behaviors, Institutions and Organizations Across Nations},
  author    = {Hofstede, Geert},
  edition   = {2},
  year      = {2001},
  publisher = {Sage Publications},
  address   = {Thousand Oaks, CA}
}

@misc{abcnews2020distress,
  author       = {Levine, Mike},
  title        = {Calls to {US} Helpline Jump 891\%, as White House Is Warned of Mental Health Crisis},
  year         = {2020},
  month        = apr,
  day          = {7},
  howpublished = {ABC News},
  url          = {https://abcnews.go.com/Politics/calls-us-helpline-jump-891-white-house-warned/story?id=70010113},
  note         = {Accessed May 26, 2026}
}

@article{matsumoto2008mapping,
  title     = {Mapping Expressive Differences Around the World: The Relationship Between Emotional Display Rules and Individualism Versus Collectivism},
  author    = {Matsumoto, David and Yoo, Seung Hee and Fontaine, Johnny and others},
  journal   = {Journal of Cross-Cultural Psychology},
  volume    = {39},
  number    = {1},
  pages     = {55--74},
  year      = {2008},
  doi       = {10.1177/0022022107311854},
  publisher = {Sage Publications}
}

@book{lipset1990continental,
  title     = {Continental Divide: The Values and Institutions of the United States and Canada},
  author    = {Lipset, Seymour Martin},
  year      = {1990},
  publisher = {Routledge},
  address   = {New York, NY}
}

@book{adams2003fire,
  title     = {Fire and Ice: The United States, Canada and the Myth of Converging Values},
  author    = {Adams, Michael},
  year      = {2003},
  publisher = {Penguin Canada},
  address   = {Toronto, ON}
}

@article{mccrae2005personalityprofiles,
  author  = {McCrae, Robert R. and Terracciano, Antonio and {{79 Members of the Personality Profiles of Cultures Project}}},
  title   = {Personality Profiles of Cultures: Aggregate Personality Traits},
  journal = {Journal of Personality and Social Psychology},
  volume  = {89},
  number  = {3},
  pages   = {407--425},
  year    = {2005},
  doi     = {10.1037/0022-3514.89.3.407}
}

@book{kahneman2011thinking,
  title     = {Thinking, Fast and Slow},
  author    = {Kahneman, Daniel},
  year      = {2011},
  publisher = {Farrar, Straus and Giroux},
  address   = {New York, NY}
}

@article{evans2008dual,
  title   = {Dual-Processing Accounts of Reasoning, Judgment, and Social Cognition},
  author  = {Evans, Jonathan St. B. T.},
  journal = {Annual Review of Psychology},
  volume  = {59},
  pages   = {255--278},
  year    = {2008},
  doi     = {10.1146/annurev.psych.59.103006.093629}
}

@article{stanovich2000individual,
  title   = {Individual Differences in Reasoning: Implications for the Rationality Debate?},
  author  = {Stanovich, Keith E. and West, Richard F.},
  journal = {Behavioral and Brain Sciences},
  volume  = {23},
  number  = {5},
  pages   = {645--665},
  year    = {2000},
  doi     = {10.1017/S0140525X00003435}
}

@article{weidinger2021ethical,
  title         = {Ethical and Social Risks of Harm from Language Models},
  author        = {Weidinger, Laura and Mellor, John and Rauh, Maribeth and Griffin, Conor and Uesato, Jonathan and Huang, Po-Sen and Cheng, Myra and Glaese, Mia and Balle, Borja and Kasirzadeh, Atoosa and Kenton, Zac and Brown, Sasha and Hawkins, Will and Stepleton, Tom and Biles, Courtney and Birhane, Abeba and Haas, Julia and Rimell, Laura and Hendricks, Lisa Anne and Isaac, William and Legassick, Sean and Irving, Geoffrey and Gabriel, Iason},
  journal       = {arXiv preprint arXiv:2112.04359},
  year          = {2021},
  eprint        = {2112.04359},
  archivePrefix = {arXiv},
  primaryClass  = {cs.CL},
  doi           = {10.48550/arXiv.2112.04359},
  url           = {https://arxiv.org/abs/2112.04359}
}

@inproceedings{bender2021dangers,
  title     = {On the Dangers of Stochastic Parrots: Can Language Models Be Too Big?},
  author    = {Bender, Emily M. and Gebru, Timnit and McMillan-Major, Angelina and Shmitchell, Shmargaret},
  booktitle = {Proceedings of the 2021 {ACM} Conference on Fairness, Accountability, and Transparency},
  series    = {{FAccT} '21},
  pages     = {610--623},
  year      = {2021},
  publisher = {Association for Computing Machinery},
  address   = {New York, NY, USA},
  doi       = {10.1145/3442188.3445922}
}

@article{wrosch2017anger_sadness_control,
  title   = {Age-Related Changes in Older Adults' Anger and Sadness: The Role of Perceived Control},
  author  = {Wrosch, Carsten and Barlow, Meaghan A. and Kunzmann, Ute},
  journal = {Psychology and Aging},
  volume  = {33},
  number  = {2},
  pages   = {350--360},
  year    = {2018},
  doi     = {10.1037/pag0000229}
}

@article{tsai2000emotionaging,
  title   = {Autonomic, Subjective, and Expressive Responses to Emotional Films in Older and Younger Chinese Americans and European Americans},
  author  = {Tsai, Jeanne L. and Levenson, Robert W. and Carstensen, Laura L.},
  journal = {Psychology and Aging},
  volume  = {15},
  number  = {4},
  pages   = {684--693},
  year    = {2000},
  doi     = {10.1037/0882-7974.15.4.684}
}

@article{li2023emotionalstimuli,
  title         = {Large Language Models Understand and Can Be Enhanced by Emotional Stimuli},
  author        = {Li, Cheng and Wang, Jindong and Zhang, Yixuan and Zhu, Kaijie and Hou, Wenxin and Lian, Jianxun and Luo, Fang and Yang, Qiang and Xie, Xing},
  journal       = {arXiv preprint arXiv:2307.11760},
  year          = {2023},
  eprint        = {2307.11760},
  archivePrefix = {arXiv},
  primaryClass  = {cs.CL},
  doi           = {10.48550/arXiv.2307.11760},
  url           = {https://arxiv.org/abs/2307.11760}
}

@misc{socialscienceone2023urls,
  author       = {{Social Science One}},
  title        = {{Facebook URL Shares Dataset}},
  year         = {2023},
  howpublished = {Dataset accessed through the Meta Researcher Platform under the applicable Research Data Agreement},
  url          = {https://socialscience.one/request-proposals-facebook-urls-dataset},
  note         = {Accessed through the Meta Researcher Platform under the applicable Research Data Agreement (RDA); accessed May 26, 2026}
}

@article{van2020using,
  title={Using social and behavioural science to support COVID-19 pandemic response},
  author={Van Bavel, Jay J. and Baicker, Katherine and Boggio, Paulo S. and others},
  journal={Nature Human Behaviour},
  volume={4},
  number={5},
  pages={460--471},
  year={2020},
  publisher={Nature Publishing Group}
}

@article{ye2026stop,
  title={Stop Drawing Scientific Claims from LLM Social Simulations Without Robustness Audits},
  author={Ye, Jinyi and Cao, Lei and Chen, Ding and Ferrara, Emilio},
  journal={arXiv preprint arXiv:2605.18890},
  year={2026}
}

@inproceedings{aher2023using,
  title={Using large language models to simulate multiple humans and replicate human subject studies},
  author={Aher, Gati V and Arriaga, Rosa I and Kalai, Adam Tauman},
  booktitle={International conference on machine learning},
  pages={337--371},
  year={2023},
  organization={PMLR}
}

@inproceedings{reimers2019sentence,
  title={Sentence-BERT: Sentence Embeddings using Siamese BERT-Networks},
  author={Reimers, Nils and Gurevych, Iryna},
  booktitle={Proceedings of the 2019 Conference on Empirical Methods in Natural Language Processing (EMNLP-IJCNLP)},
  pages={3982--3992},
  year={2019},
  publisher={Association for Computational Linguistics}
}

\appendix

\section{Extended Persona Dataset Validation and Calibration}
\label{appendix:validation}

This appendix provides detailed validation results for the LLM-generated persona dataset, including institutional community balance, moral foundations assignment verification, psychological trait calibration, and linguistic response structure metrics. These analyses supplement the high-level findings summarized in the Results Section.

\subsection{Moral Foundations Assignment Verification}
\label{app:moral}
 
Moral foundation scores were assigned from ideological category using a categorical lookup table with limited individual variance. Each persona inherited the ideology label from their institutional community and received base moral foundation values, with small perturbations ($\sigma^2 \approx 0.01$) to introduce within-group variation while preserving ideological coherence.
 
The assignment uses five discrete base levels (Very Low = 0.10, Low = 0.30, Medium = 0.50, High = 0.70, Very High = 0.90) with constrained noise ensuring 95\% of values remain within $\pm$10\% of the scale range. Table~\ref{tab:moral_foundations_political} demonstrates two key patterns: (1) substantial overlap among centrist groups (Center Left, Center, Center Right) across most moral foundations, and (2) clear separation between ideological extremes (Left vs.\ Right) on binding foundations (Authority, Sanctity). This is an implementation verification check. Behavioral validation occurs through event-conditioned emotional response analysis (Section~\ref{sec:results}).
 
\begin{table}[htbp] 
\centering
\caption{Moral foundations assignment verification by political orientation.}
\label{tab:moral_foundations_political}
\footnotesize
\begin{threeparttable}
\begin{tabular}{lrrrrrr}
\toprule
\textbf{Political Group} & \textbf{Care} & \textbf{Fairness} & \textbf{Loyalty} & \textbf{Authority} & \textbf{Sanctity} & \textbf{Liberty} \\
\midrule
Left (n=139) & 0.89 & 0.89 & 0.50 & 0.50 & 0.31 & 0.70 \\
Center Left (n=150) & 0.72 & 0.72 & 0.51 & 0.52 & 0.48 & 0.69 \\
Center (n=78) & 0.71 & 0.71 & 0.51 & 0.52 & 0.49 & 0.69 \\
Center Right (n=124) & 0.70 & 0.70 & 0.68 & 0.88 & 0.88 & 0.52 \\
Right (n=46) & 0.70 & 0.70 & 0.68 & 0.86 & 0.86 & 0.52 \\
\midrule
\multicolumn{7}{l}{\textit{\textbf{Centrist Overlap Demonstration}}} \\
Center Left $\cap$ Center & \checkmark & \checkmark & \checkmark & \checkmark & \checkmark & \checkmark \\
Center $\cap$ Center Right & \checkmark & \checkmark & \checkmark & \checkmark & \checkmark & \checkmark \\
Combined Range (CL--C--CR) & 0.70--0.72 & 0.70--0.72 & 0.51--0.68 & 0.52--0.88 & 0.48--0.88 & 0.52--0.69 \\
\midrule
\multicolumn{7}{l}{\textit{\textbf{Extreme Separation (Left vs.\ Right)}}} \\
$\Delta$ (L--R) & +0.19 & +0.19 & --0.18 & --0.36 & --0.55 & +0.18 \\
Separation Ratio$^a$ & 0.24 & 0.24 & 0.23 & 0.45 & 0.69 & 0.23 \\
Cohen's $d^b$ & 2.89 & 2.89 & --2.71 & --3.45 & --4.22 & 2.71 \\
\bottomrule
\end{tabular}
\begin{tablenotes}
\footnotesize
\item[$^a$] Separation ratio = $|\Delta|$ / scale range (0.80). Values $>$0.5 indicate strong separation.
\item[$^b$] Cohen's $d$ calculated with within-group variance ($\sigma^2 \approx 0.01$). Moderate-to-large effect sizes reflect categorical base values with limited noise. Centrist groups show substantial distributional overlap (checkmarks indicate overlapping ranges), while extreme groups separate on binding foundations.
\end{tablenotes}
\end{threeparttable}
\end{table}

\subsection{OCEAN Personality Trait Calibration}

To improve consistency with reported cross-cultural personality distributions, personas were assigned OCEAN (Big Five) personality traits drawn from normal distributions parameterized by cross-cultural means from \cite{schmitt2007geographic, mccrae2005personalityprofiles}. Table~\ref{tab:ocean_validation} reports LLM-generated averages against empirical benchmarks for both the U.S.\ and Canada.

\begin{table}[htbp]
\centering
\caption{OCEAN personality trait validation: LLM-generated vs.\ empirical research data.}
\label{tab:ocean_validation}
\footnotesize
\begin{threeparttable}
\begin{tabular}{p{0.2\linewidth}rrrrrr}
\toprule
\multirow{2}{*}{\textbf{Trait}} &
\multicolumn{2}{c}{\textbf{U.S.}} &
\multicolumn{2}{c}{\textbf{Canada}} &
\multicolumn{2}{c}{\textbf{Deviation}$^a$} \\
\cmidrule(lr){2-3} \cmidrule(lr){4-5} \cmidrule(lr){6-7}
 & LLM & Obs. & LLM & Obs. & U.S. & CA \\
\midrule
Openness          & 0.492 & 0.500 & 0.485 & 0.487 & -0.008 & -0.002 \\
Conscientiousness & 0.495 & 0.500 & 0.486 & 0.490 & -0.005 & -0.004 \\
Extraversion      & 0.493 & 0.500 & 0.490 & 0.483 & -0.007 & +0.007 \\
Agreeableness     & 0.500 & 0.500 & 0.501 & 0.491 &  0.000 & +0.010 \\
Neuroticism       & 0.486 & 0.500 & 0.502 & 0.506 & -0.014 & -0.004 \\
\midrule
\textbf{Mean Absolute Deviation}
& \multicolumn{2}{c}{0.007}
& \multicolumn{2}{c}{0.005}
& \multicolumn{2}{c}{} \\
\bottomrule
\end{tabular}
\begin{tablenotes}
\footnotesize
\item[$^a$] Deviations are computed as LLM mean minus empirical benchmark.
\end{tablenotes}
\end{threeparttable}
\end{table}
 
Trait alignment demonstrated strong psychological validity. Canadian personas scored slightly higher in \textit{Agreeableness} and lower in \textit{Openness}, reflecting cultural values emphasizing modesty and cooperation. Institutional variation followed logical patterns: governmental personas showed highest Openness ($\mu=0.504$), university personas lowest Conscientiousness ($\mu=0.479$), and industry personas highest Agreeableness ($\mu=0.508$).

Demographic effects were subtle but interpretable---age correlated positively with Openness ($r=+0.089$) and Extraversion ($r=+0.071$), and negatively with Agreeableness ($r=-0.060$) and Neuroticism ($r=-0.069$). Gender effects were minimal (all $|\Delta|<0.010$), indicating consistent calibration across demographic groups.

\begin{figure}[htbp]
\centering
\includegraphics[width=\columnwidth]{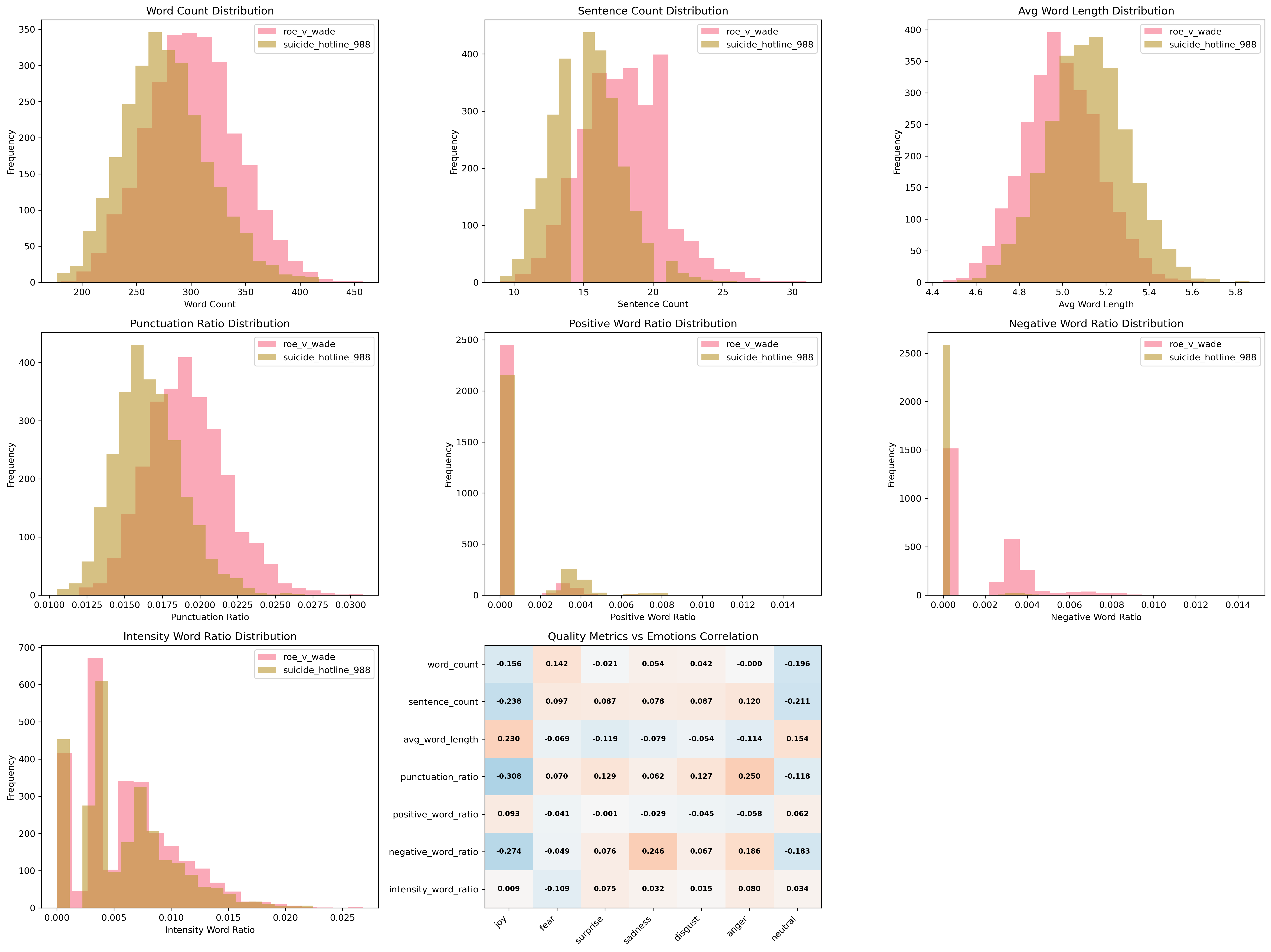}
\caption{Response structure metrics across Roe v.\ Wade and Suicide Hotline~988 responses. Shown distributions include word count, sentence count, average word length, punctuation ratio, and affective word ratios.}
\label{fig:response_quality_appendix}
    \Description{Panel of distribution plots for response structure metrics
      computed over Roe v. Wade and Suicide Hotline 988 responses: word count,
      sentence count, average word length, punctuation ratio, and affective
      word ratios.}
\end{figure}

\subsection{Response Structure and Emotional Intensity Distributions}

Finally, the linguistic realism of persona responses was assessed through word-level and sentence-level structure metrics. Figure~\ref{fig:response_quality_appendix} visualizes distributions for word count, sentence count, average word length, punctuation ratio, and affective word proportions (positive, negative, intensity).

Roe v.\ Wade and Suicide Hotline~988 responses exhibited comparable response lengths, but differed in affective tone. Roe v.\ Wade posts contained higher negative word ratios, corresponding to elevated sadness ($r=0.246$) and anger ($r=0.186$). In contrast, Suicide Hotline~988 posts contained more positive and high-intensity words, correlating with joy ($r=0.295$). These structural measures provided an independent validation of event-conditioned emotional-response realism across generated responses.

 \subsection{Prompt Construction}
\label{appendix:prompts}

Persona prompts are assembled from four modular components:

\begin{enumerate}
    \item \textbf{Persona Identity}: name, age, gender, education, occupation, institutional affiliation, country, and political leaning.
    \item \textbf{Psychological Profile}: OCEAN personality traits, moral foundation scores, and baseline or event-conditioned emotion cues, depending on the experimental condition.
    \item \textbf{Event Context}: event title and description, retrieved news snippets, persona stance, and event salience.
    \item \textbf{Conversational Context}: recent posts from neighboring personas in the social network, ordered chronologically.
\end{enumerate}

Table~\ref{tab:prompt_conditions_appendix} summarizes the prompt components included in each experimental condition.

\begin{table}[htbp]
\centering
\caption{Prompt components by experimental condition.}
\label{tab:prompt_conditions_appendix}
\scriptsize
\setlength{\tabcolsep}{3pt}
\begin{tabularx}{\columnwidth}{l*{5}{>{\centering\arraybackslash}X}}
\toprule
\textbf{Component} & \textbf{D} & \textbf{D+B} & \textbf{D+B+M} & \textbf{D+BMO} & \textbf{DMOE} \\
\midrule
Persona Identity      & \checkmark & \checkmark & \checkmark & \checkmark & \checkmark \\
OCEAN Traits          & -- & -- & -- & \checkmark & \checkmark \\
Moral Foundations     & -- & -- & \checkmark & \checkmark & \checkmark \\
Baseline Emotions     & -- & \checkmark & \checkmark & \checkmark & -- \\
Event Emo.            & -- & -- & -- & -- & \checkmark \\
Event Context         & \checkmark & \checkmark & \checkmark & \checkmark & \checkmark \\
Conversation          & \checkmark & \checkmark & \checkmark & \checkmark & \checkmark \\
\bottomrule
\end{tabularx}
\end{table}

All prompts conclude with an instruction directing the LLM to generate a social media post consistent with the persona, event, and conversational context. Responses are generated using GPT-4o-mini and automatically validated for length, coherence, persona consistency, and schema compliance before downstream analysis.

\section{COVID-19 Vaccine Emotional Response Validation}
\label{app:covid_validation}

We further validate synthetic persona plausibility using COVID-19 vaccine discourse, comparing emotional reactions to vaccine-effectiveness and vaccine-harm narratives against Facebook URL Shares data. 

\begin{figure}[htbp]
    \centering
    \includegraphics[width=\textwidth]{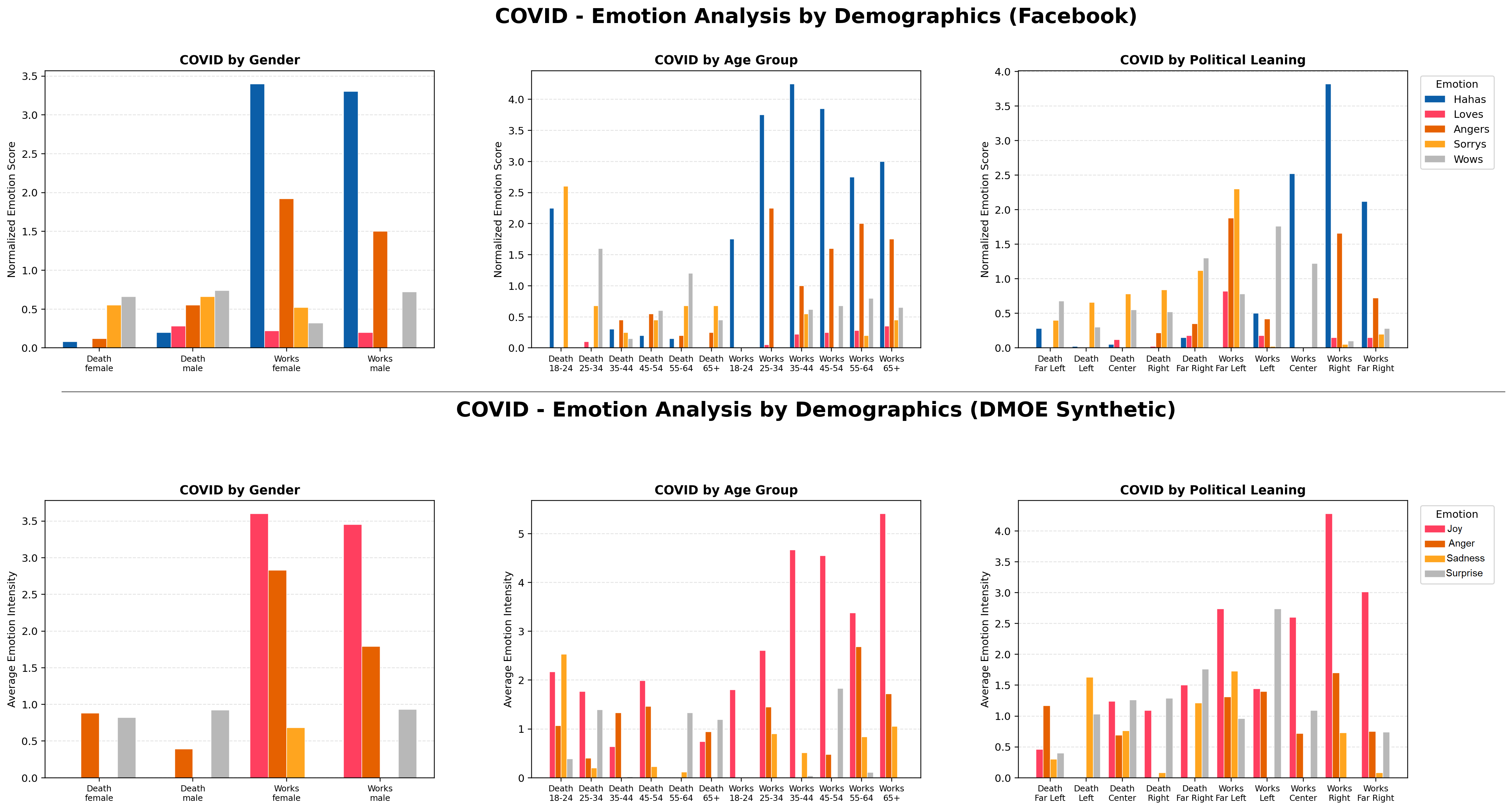}
    \caption{Comparison of Facebook COVID-19 emotional reactions and D+M+O+E synthetic persona emotional-response patterns across gender, age, and political-orientation demographics.}
    \label{fig:covid_combined}
    \Description{Two rows of grouped bar charts comparing Facebook COVID-19
      reaction shares with synthetic persona emotional response shares under
      the D+M+O+E condition, each broken down by gender, age group, and
      political orientation.}
\end{figure}

Figure~\ref{fig:covid_combined} reveals several important similarities and differences between the Facebook reactions and the synthetic personas across vaccine-supportive (“vaccines work”) and vaccine-harm (“vaccines death”) narratives. In both datasets, vaccine-supportive narratives consistently generated stronger positive emotional engagement, particularly through high levels of joy/love reactions and elevated anger toward opposing viewpoints. In contrast, vaccine-harm narratives produced comparatively weaker positive engagement and stronger sadness- and surprise-oriented reactions, suggesting greater uncertainty, distrust, and emotional caution.

The Facebook data showed substantially sharper emotional polarization between ideological groups than the synthetic personas. For vaccine-supportive narratives, politically right-leaning Facebook users exhibited extremely high positive emotional intensity alongside elevated anger responses, indicating both strong support and adversarial engagement toward competing claims. The synthetic personas reproduced this directional ideological split, but with smoother and less extreme emotional magnitudes. This suggests that the hierarchical prompting framework successfully captured relative ideological positioning while moderating the intensity of highly polarized online behavior.

Differences also emerged in how emotional reactions were distributed across age groups. In the Facebook data, younger cohorts displayed stronger high-arousal responses under both vaccine-supportive and vaccine-harm narratives, particularly anger and joy/love reactions. Older cohorts showed comparatively more stable and lower-intensity emotional patterns. The synthetic personas reproduced these broad age-conditioned trends, although the emotional transitions between age groups appeared more gradual and less noisy than in the real-world data. This smoothing effect likely reflects the tendency of LLM-generated personas to produce more internally coherent emotional outputs than naturally occurring social-media behavior.

Gender patterns were comparatively stable across both datasets, with one notable difference involving sadness expression. Female groups generally exhibited slightly higher sadness- or sorry-oriented responses, particularly under vaccine-harm narratives and in the vaccine-supportive condition, while male groups showed comparatively lower sadness expression. This under-expression of sadness among male groups is important because it appears in both the Facebook reactions and the LLM-generated synthetic personas, suggesting that the model captured a gender-conditioned emotional tendency rather than only aggregate sentiment. Under vaccine-supportive narratives, both male and female groups showed elevated positive engagement, although male groups in the Facebook data demonstrated somewhat stronger anger responses toward opposing claims. The synthetic personas preserved these directional differences while again producing less extreme emotional variance overall.

A key distinction between the two narrative conditions is that vaccine-supportive discourse generated emotionally amplified and polarized engagement, whereas vaccine-harm discourse produced more mixed and uncertainty-oriented reactions. In both the Facebook and synthetic datasets, vaccine-harm narratives were associated with higher sadness and surprise levels and weaker positive emotional intensity. This suggests that emotionally charged public-health discourse does not simply divide populations into positive versus negative reactions, but instead produces distinct emotional structures depending on whether the discourse centers on reassurance, institutional trust, perceived danger, or uncertainty.

Importantly, the synthetic personas did not merely reproduce average sentiment levels. Instead, they preserved the relative structure of demographic and ideological emotional variation across both narrative conditions, including the gendered under-expression of sadness among male groups. While the synthetic outputs remained smoother and less extreme than the Facebook reactions, the framework reproduced several broader polarization patterns, demographic modulation, and emotional differentiation associated with prolonged COVID-19 vaccine discourse.

\section{Internal Consistency Validation}
\label{appendix:internal_consistency}

 This appendix evaluates whether the synthetic persona generation framework produces personas that are internally consistent with the intended design. Specifically, we assess demographic fidelity, community-level consistency, opinion generation, and schema integrity. These evaluations verify implementation correctness rather than external behavioral realism. 

\subsection{Demographic Alignment}

 Demographic distributions of generated personas were compared with reference statistics using the \textbf{Jensen--Shannon Divergence (JSD)}, a symmetric measure of similarity between probability distributions. Lower values indicate closer agreement, with 0 representing identical distributions. 

\begin{table}[htbp]
\centering
\caption{Demographic Alignment with Reference Distributions} 
\label{tab:demographic_alignment}
\small
\begin{tabular}{lccc}
\toprule
\textbf{Country} & \textbf{Variable} & \textbf{JSD} & \textbf{Interpretation} \\
\midrule
Canada & Gender & \stat{0.006} & Excellent agreement \\
Canada & Age Distribution & \stat{0.028} & High agreement \\
United States & Gender & \stat{0.008} & Excellent agreement \\
United States & Age Distribution & \stat{0.021} & High agreement \\
\bottomrule
\end{tabular}
\end{table}

 Overall, demographic distributions closely matched their reference populations, with all Jensen--Shannon divergence values below \stat{0.05}, indicating minimal distributional differences.

\subsection{Community Consistency}

To verify that personas inherited characteristics from their assigned communities while retaining individual variability, we computed the \textbf{Intraclass Correlation Coefficient (ICC)}. Higher ICC values indicate stronger within-community similarity.

\begin{table}[htbp]
\centering
 \caption{Community-Level Consistency}
\label{tab:community_consistency}
\small
\begin{tabular}{lcc}
\toprule
\textbf{Attribute} & \textbf{ICC} & \textbf{Interpretation} \\
\midrule
Political Leaning & \stat{0.91} & Strong consistency \\
Age & \stat{0.83} & Strong consistency \\
Education Level & \stat{0.86} & Strong consistency \\
\bottomrule
\end{tabular}
\end{table}

All attributes exhibited ICC values above \stat{0.80}, indicating that personas consistently inherited community-level characteristics while preserving individual variation.
 
\subsection{Opinion Generation Consistency}

The framework assumes that higher topic salience produces lower opinion variability. This relationship was evaluated using the \textbf{Pearson correlation coefficient} between topic salience and observed opinion variance.

\begin{table}[htbp]
\centering
\caption{Salience and Opinion Consistency}
\label{tab:salience_consistency}
\small
\begin{tabular}{lcc}
\toprule
\textbf{Metric} & \textbf{Value} & \textbf{Interpretation} \\
\midrule
Pearson Correlation ($r$) & \stat{-0.84} & Strong negative relationship \\
Coefficient of Determination ($R^2$) & \stat{0.71} & High explanatory power \\
\bottomrule
\end{tabular}
\end{table}

The strong negative correlation (\stat{$r=-0.84$}) confirms that higher topic salience consistently resulted in lower opinion variance, matching the intended behavior of the generation framework.

\end{document}